\documentclass{article}

\PassOptionsToPackage{numbers, sort}{natbib}

\usepackage[preprint]{neurips_2023}

\usepackage[utf8]{inputenc} 
\usepackage[T1]{fontenc}    
\usepackage{url}            
\usepackage{booktabs}       
\usepackage{amsfonts}       
\usepackage{nicefrac}       
\usepackage{microtype}      
\usepackage{xcolor,colortbl}       
\usepackage{comment}
\usepackage{adjustbox}
\usepackage{algpseudocode}
\usepackage{algorithm}
\usepackage{graphicx}
\usepackage{amsmath}
\usepackage{amssymb}
\usepackage{wrapfig}
\usepackage{comment}
\usepackage{shortbold}

\usepackage[pagebackref,breaklinks,colorlinks]{hyperref}
\usepackage{xspace}

\usepackage[capitalize]{cleveref}
\crefname{section}{Sec.}{Secs.}
\Crefname{section}{Section}{Sections}
\Crefname{table}{Table}{Tables}
\crefname{table}{Tab.}{Tabs.}
\Crefname{equation}{Equation}{Equations}
\crefname{equation}{Eq.}{Eqs.}
\Crefname{figure}{Figure}{Figures}
\crefname{figure}{Fig.}{Figs.}
\Crefname{table}{Table}{Tables}
\crefname{table}{Tab.}{Tabs.}

\usepackage{authblk}
\usepackage{bbm}

\title{NIFTY: Neural Object Interaction Fields \\ for Guided Human Motion Synthesis}

\makeatletter \renewcommand\AB@affilsepx{\quad \protect\Affilfont} \makeatother

\author[1,2]{ Nilesh Kulkarni}
\author[ {~~}3]{ \qquad \qquad Davis Rempe\thanks{Work done while at Stanford University}}
\author[2]{ \qquad \qquad Kyle Genova}
\author[2]{\qquad \qquad Abhijit Kundu}
\author[2]{Justin Johnson}
\author[2]{\qquad  David Fouhey}
\author[2,4]{\qquad Leonidas Guibas}
\affil[1]{University of Michigan}
\affil[2]{Google}
\affil[3]{NVIDIA}
\affil[4]{Stanford University}

\newcommand{\normalN}[0]{\mathcal{N}}
\newcommand{\traj}[0]{\boldsymbol{\tau}}

\newcommand{\ours}[0]{NIFTY\xspace}
\newcommand{\humanise}[0]{Cond.\ VAE~\cite{wang2022humanise}\xspace}
\newcommand{\mdmOnly}[0]{Cond.\ MDM~\cite{tevet2022human}\xspace}

\newcommand{\parnobf}[1]{\vspace{-0.7mm} \par \noindent {\bf {#1}.}}

\usepackage{amsmath,amsfonts,bm}

\def\eqref#1{equation~\ref{#1}}

\def\1{\bm{1}}

\newcommand{\test}{\mathcal{D_{\mathrm{test}}}}

\def\eps{{\epsilon}}

\def\vj{{\bm{j}}}

\def\vt{{\bm{t}}}

\DeclareMathAlphabet{\mathsfit}{\encodingdefault}{\sfdefault}{m}{sl}
\SetMathAlphabet{\mathsfit}{bold}{\encodingdefault}{\sfdefault}{bx}{n}

\def\sR{{\mathbb{R}}}

\newcommand\norm[1]{\lVert#1\rVert}

\newcommand{\ie}{\textit{i}.\textit{e}., }
\newcommand{\eg}{\textit{e}.\textit{g}.}
\newcommand{\etc}{\textit{etc.}}

\newcommand{\davis}[1]{\textcolor{red}{Davis: #1}}

\definecolor{lastblue}{HTML}{5F5F72}
\definecolor{lightpink}{HTML}{ffb6c1}
\definecolor{ibm1}{HTML}{648FFF}
\definecolor{ibm2}{HTML}{DC267F}
\definecolor{ibm3}{HTML}{FE6100}
\definecolor{ibm4}{HTML}{FFB000}
\definecolor{ibm5}{HTML}{785EF0}
\definecolor{ibm6}{HTML}{88CCEE}

\definecolor{dmcolor}{RGB}{157, 195, 230}
\definecolor{oifcolor}{RGB}{169, 209, 142}
\definecolor{ibm_red}{HTML}{E62325}
\definecolor{ibm_green}{HTML}{24A148}
\definecolor{ibm_yellow}{HTML}{f1c21b}
\definecolor{ibm_purple}{HTML}{C22DD5}
\definecolor{ibm_teal}{HTML}{009d9a}
\definecolor{ibm_ultramarine}{HTML}{648FFF}
\definecolor{ibm_blue}{HTML}{002d9c}

\definecolor{grey}{HTML}{808080}
\definecolor{grey}{HTML}{808080}
\definecolor{ibm_red80}{HTML}{750e13}
\definecolor{orange}{HTML}{FFA500}
\definecolor{blue}{HTML}{0000FF}

\definecolor{dred}{RGB}{242, 220, 219}
\definecolor{dblue}{RGB}{220, 230, 242}
\definecolor{dpurple}{RGB}{235, 212, 225}

\newcolumntype{s}{>{\columncolor{dred}}c}
\newcolumntype{r}{>{\columncolor{dblue}}c}

\begin{document}

\maketitle

\begin{abstract}
We address the problem of generating realistic 3D motions of humans interacting with objects in a scene. Our key idea is to create a neural interaction field attached to a specific object, which outputs the distance to the valid interaction manifold given a human pose as input. This interaction field guides the sampling of an object-conditioned human motion diffusion model, so as to encourage plausible contacts and affordance semantics. To support interactions with scarcely available data, we propose an automated synthetic data pipeline. For this, we seed a pre-trained motion model, which has priors for the basics of human movement, with interaction-specific anchor poses extracted from limited motion capture data. Using our guided diffusion model trained on generated synthetic data, we synthesize realistic motions for sitting and lifting with several objects, outperforming alternative approaches in terms of motion quality and successful action completion. We call our framework \textbf{\ours}: \textbf{N}eural \textbf{I}nteraction \textbf{F}ields for \textbf{T}rajectory s\textbf{Y}nthesis.
Project Page: \href{nileshkulkarni.github.io/nifty}{\texttt{https://nileshkulkarni.github.io/nifty}}
\end{abstract}

\section{Introduction}

Animating a character to sit in a chair or pick up a box is useful in gaming, character animation, and populating digital twins. Yet, generating realistic 3D human motion trajectories with objects is challenging for two main reasons.
One challenge is creating \textit{effective models} that capture the nuance of human movements, particularly during the final phase of object interaction called the "last mile." Unlike navigation that is primarily collision avoidance, the last mile involves intricate contacts and object affordances, which influence the motion. 
The second challenge is \textit{acquiring paired data} that includes high-quality human motions and diverse object shapes, which is essential for training. 
\begin{figure}[t]
    \centering
    \noindent
    \begin{adjustbox}{max width=\linewidth}
    \begin{tabular}{c}
    \noindent\includegraphics{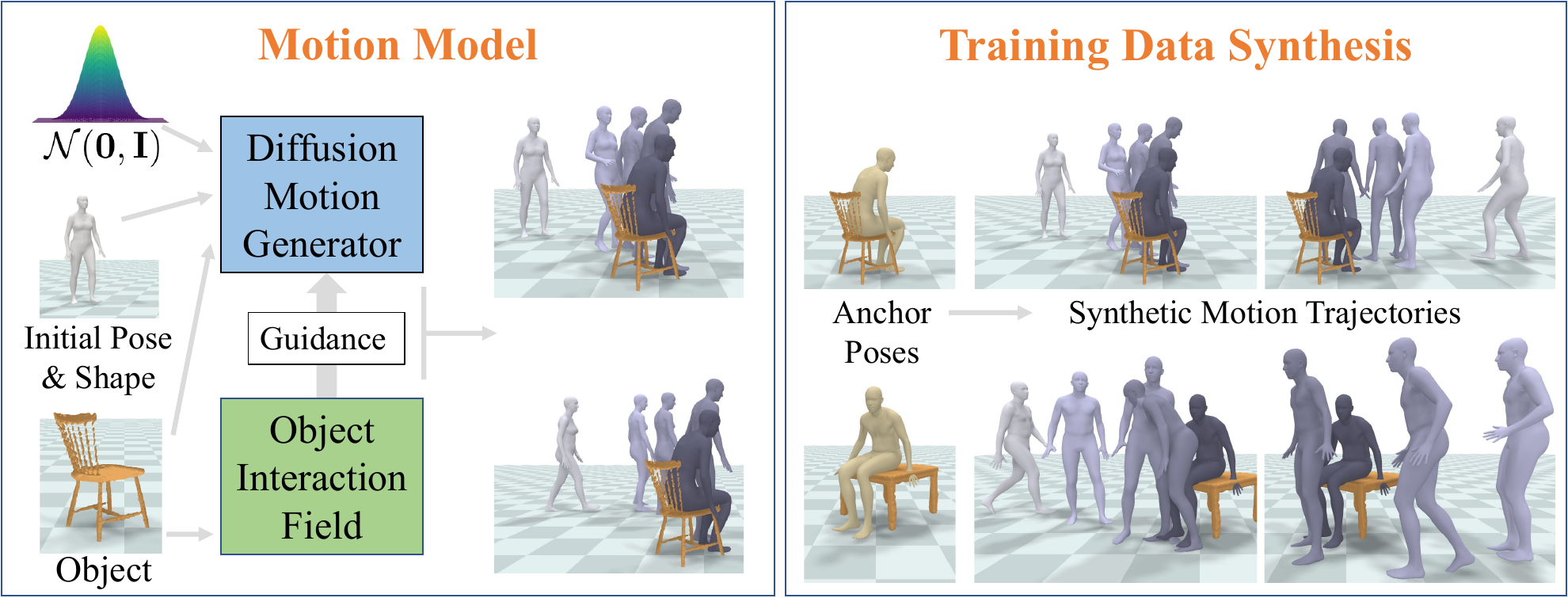} 
    \end{tabular}
    \end{adjustbox}
    \caption{\textbf{\ours Overview.} (Left) Our learned \textcolor{oifcolor}{\textbf{object interaction field}} guides an object-conditioned \textcolor{dmcolor}{\textbf{diffusion model}} during sampling to generate plausible human-object interactions like sitting. (Right) Our automated \textit{training data synthesis pipeline} generates data for this model by combining a scene-unaware motion model with small quantities of annotated interaction anchor pose data.}
    \label{fig:overview}
    \vspace{-5mm}
\end{figure}

Recent approaches to motion modeling can synthesize realistic human movements using state-of-the-art generative models~\cite{tevet2022human,rempe2021humor,henter2020moglow}. They are, however, scene-agnostic and cannot produce interactions with specific objects.

To address this, some approaches condition motion synthesis on scene geometry (\eg, a scanned point cloud)~\cite{wang2021synthesizing,wang2022towards,wang2022humanise,huang2023diffusion}. This enables learning object interactions, but motion quality is hindered by the lack of paired full scene-motion data. 
Other approaches~\cite{zhang2022couch,hassan2021stochastic,starke2019neural} instead focus on a small set of interactions with a single type of object (\eg, sitting in a chair), and can produce high-quality motions in their domain. However, these methods require a high-quality motion capture (mocap) dataset for each action/object separately and may make action-specific modeling assumptions (\eg, affecting the human approach and/or contact with object).

In this work, we tackle both the \textit{modeling} and \textit{data} aspects of interaction synthesis to enable generating realistic interactions with a variety of objects, such as sitting on a chair, table, or stool and lifting a suitcase, chair, \etc.
We extend a human motion diffusion model~\cite{tevet2022human} to condition on object geometry, and pair it with a \textit{learned object interaction field} to encourage realistic movements at test time in the last mile of interaction. 
To train this model and overcome the lack of available mocap data, we develop an \textit{automated data pipeline} that leverages a powerful pre-trained and scene-agnostic human motion model. 
As shown in \cref{fig:overview}, our interaction field, diffusion model, and motion data pipeline make up a general framework to synthesize human-object interactions for a desired character that is flexible to multiple actions, even when dense mocap data is {\it unavailable}. 
We refer to this framework as \textbf{\ours}: \textbf{N}eural \textbf{I}nteraction \textbf{F}ields for \textbf{T}rajectory s\textbf{Y}nthesis.

To ensure realistic motions in the last mile of interaction, we propose an object-centric interaction field that takes in a human pose and learns to regress the distance to a valid interaction pose (\eg, the final sitting pose). 
At test time, our object-conditioned diffusion model is \textit{guided} by this interaction field to encourage high-quality motions. 
Unlike manually designed guidance objectives that encourage contact and discourage penetration~\cite{huang2023diffusion}, our interaction field is data-driven and implicitly captures notions of contact, object penetration, and any other factors learned from data. 

We propose using synthetic data generation to enable learning interactions from limited mocap data.
In particular, we leverage a pre-trained motion model~\cite{rempe2021humor} that produces high-quality motions but is unaware of object geometry. 
Starting from an anchor pose that captures the core of a desired interaction (\eg, the final sitting pose in \cref{fig:overview}, right), the pre-trained model is used to sample a large variety of motions that \textit{end} in the anchor pose.
This approach generates a diverse set of plausible interactions from only a handful of anchor poses, which are readily available from existing small datasets~\cite{bhatnagar22behave} or are relatively easy to capture. 

We evaluate \ours on sitting and lifting interactions for a variety of objects, demonstrating the superior quality of synthesized human motions compared to alternative approaches.
Overall, this work contributes \textbf{(1)} a novel object interaction field approach that guides an object-conditioned human motion diffusion model to synthesize realistic interactions, \textbf{(2)} an automated synthetic data generation technique to produce large numbers of plausible interactions from limited pose data, and \textbf{(3)} high-quality motion synthesis results for human interactions with several objects.
\section{Related Work}

\parnobf{Synthesizing Human Motion and Interactions} 
While various methods have been successful in generating human motion in isolation ~\cite{rempe2021humor,henter2020moglow,tevet2022human,holden2017phase,zhang2022wanderings,petrovich2021action}, our work is primarily focused on incorporating environmental context~\cite{corona2020context,chao2021learning}. Some approaches condition motion generation on scanned scene geometry that encompasses multiple objects~\cite{wang2021synthesizing,wang2021scene,wang2022towards,huang2023diffusion}, but these methods typically offer limited control over the specific objects for interaction.
Object-centric models are trained to generate motions for a single character~\cite{starke2019neural,hassan2021stochastic} and limited actions~\cite{zhang2022couch}, such as sitting on a chair. These models heavily rely on high-quality motion capture datasets but still exhibit issues like floating, skating, and penetration. Our work also focuses on individual objects but utilizes diffusion guidance and a learned interaction field to minimize undesired artifacts. In contrast to prior work, we train our models using a novel data generation pipeline to learn interactions from limited data. Our focus is on macroscopic interactions like sitting and lifting with objects, distinguishing us from other works that generate full-body motions for grasping and manipulation~\cite{taheri2020grab,taheri2022goal,ghosh2022imos}.

\parnobf{Motion Modeling with Diffusion} 
Following success for image~\cite{ho2020denoising,nichol2021improved} and video~\cite{ho2022video} generation, diffusion models~\cite{sohl2015deep} have shown promise in modeling motion for robots~\cite{janner2022diffuser} and pedestrians~\cite{rempe2023trace}. 
Recently, diffusion models have been successful in generating full-body 3D human motion~\cite{tevet2022human,zhang2022motiondiffuse,tseng2022edge,dabral2022mofusion}.
SceneDiffuser~\cite{huang2023diffusion} generates human motion conditioned on a point cloud from a scanned scene. It employs gradient-based guidance and analytic objectives to ensure collision-free, contact-driven, and smooth motion during the denoising process.
On the contrary, our approach is object-centric and does not rely on noisy motions~\cite{hassan2019prox} for training.
Our \textit{data-driven} interaction field guides denoising by implicitly capturing plausible interactions and obviating the need for hand-designed objectives.

\parnobf{Neural Distance Fields for Pose and Interaction} 
Neural networks have been used to learn a parametric function that outputs a distance given a query coordinate~\cite{xie2022neural}. 
Grasping Fields~\cite{karunratanakul2020grasping} parameterize hand-object grasping through a spatial field that outputs distances to valid hand-object grasps. 
Pose-NDF~\cite{tiwari2022pose} learns an object-unaware distance field in the full-body pose space for human poses. 
NGDF~\cite{weng2023ngdf} and SE(3)-DiffusionFields~\cite{urain2022se} learn a field in the robot gripper pose space to define a manifold of valid object grasps. 
Our object interaction field extends this idea to full-body human-object interactions by learning to predict the distance between a human pose and the interaction pose manifold. Unlike prior works, we use this field to guide denoising.

\parnobf{Human Interaction Data} 
Though large-scale mocap data is available to train scene-agnostic human motion models~\cite{mahmood2019amass}, learning human-object interactions is hampered by the challenge of capturing humans in scenes.
Datasets that contain full scene scans paired with human motion~\cite{savva2016pigraphs,hassan2019prox,guzov2021human,zheng2022gimo,huang2022capturing} are relatively small and often noisy due to capture difficulties.
Other datasets contain single-object interactions with a small set of objects~\cite{bhatnagar22behave,zhang2022couch,jiang2022chairs,taheri2020grab,hassan2021stochastic}. These are better quality due to simpler capture conditions, but are small with limited scope. 
Recent approaches circumvent the data issue through automated synthetic data generation. For example, 3D scenes can be inferred from pre-recorded human motions to get plausible paired scene-motion data~\cite{ye2022scene,yi2022mime,wang2022humanise}. However, motions from these methods are limited to available pre-recorded data. 
Our data generation pipeline requires only a small set of interaction anchor poses and generates novel motions not contained in prior datasets using tree-based rollouts~\cite{zhang2022wanderings} from a pre-trained generative model~\cite{rempe2021humor}.
\section{Method}
In this section, we detail our \ours pipeline for learning to synthesize realistic human-object interaction motions. 
\S\ref{subsec:diffusion} introduces a conditional diffusion model to generate human motions given the geometry of an object. 
\S\ref{subsec:oif} details the object-centric interaction field, which guides the denoising process of the diffusion model to capture the nuances of interactions in a data-driven way.
In  \S\ref{subsec:data_gen}, we discuss the  synthetic data generation using a pre-trained motion model that is seeded with anchor poses from a smaller dataset. This data is used to train the diffusion model and interaction field.

\vspace{-2mm}
\subsection{Motion Generation using Diffusion Modeling}
\label{subsec:diffusion}
\parnobf{Motion Representation} 
Motion generation is formulated as predicting a sequence of 3D human pose states that capture a person's motion over time. 
The pose state representation is based on the SMPL body model~\cite{smpl} and is similar to prior successful human motion diffusion models~\cite{tevet2022human,hm3d}.
The human pose state $X_{i}$ at frame $i$ in a motion sequence is:
\begin{equation}
    X_{i} = \{  \vj_{i}^{p}, \vj_{i}^{r}, \vj_{i}^{v}, \vj_{i}^{\omega}, \vt^{p}_{i}, \vt^{v}_{i}\},
\end{equation}
which includes joint positions $\vj_{i}^{p} \in \sR^{3 \times 22}$, rotations $\vj_{i}^{r} \in \sR^{6 \times 22}$, velocities $\vj_{i}^{v} \in \sR^{3 \times 22}$, and angular velocities $\vj_{i}^{\omega} \in \sR^{3\times 22}$ for all $22$ SMPL joints including the root (pelvis).
Additionally, the SMPL global translation $\vt_{i}^{p} \in \sR^{3}$ and velocity $\vt_{i}^{v} \in \sR^{3}$ are included. 
A motion (trajectory) is a sequence of $N$ poses denoted as $\traj = \{ X_{1}, \dots, X_{N} \}$ where all poses are in a canonical coordinate frame, namely, the local frame of the pose $X_{1}$ at the first timestep where the human is at the origin and its front-facing vector is aligned with the $+Y$ axis.

\begin{figure}[t]
    \centering
    \noindent
    \begin{adjustbox}{max width=\linewidth}
   
    \includegraphics{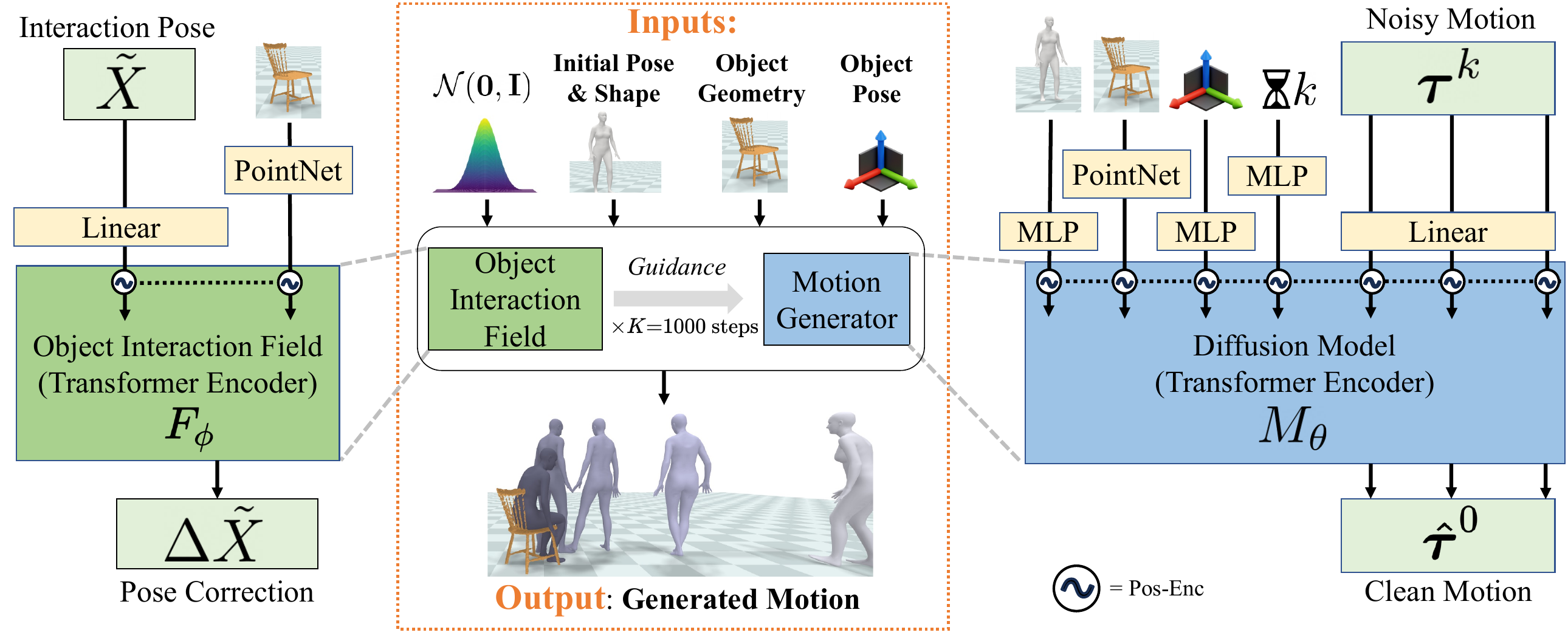}
    \end{adjustbox}
    \vspace{-1.5em}
    \caption{{\bf Model Architecture.} Our full motion synthesis method (middle) consists of an \textcolor{oifcolor}{\textbf{object interaction field}} $F_\phi$ (left), which guides the \textcolor{dmcolor}{\textbf{diffusion model}} $M_\theta$ (right) at sampling time to produce plausible interaction motions. At each step $k \in [0, K=1000]$ of denoising, the diffusion model predicts a clean motion $\hat{\traj}^{0}$ from a noisy motion input $\traj^{k}$ and conditioning information. The object interaction field takes the last pose from the diffusion output as input, and uses guidance to push the pose towards the valid interaction manifold using a predicted pose correction. 
    }
    \label{fig:method}
    \vspace{-4mm}
\end{figure}
\parnobf{Model Formulation}
The diffusion model simultaneously generates all human poses in a motion sequence~\cite{tevet2022human} to achieve a desired interaction.
Intuitively, diffusion is a noising process that converts clean data into noise. 
We want our motion model to learn the reverse of this process so that realistic motions can be generated from randomly sampled noise. 
Mathematically, forward diffusion is a Markov process with a transition probability distribution:
\begin{equation}
     q(\traj^{k}|\traj^{k-1})  := \normalN(\traj^{k}; \mu=\sqrt{1-\beta^{k}}\traj^{k-1}, \sigma=\beta^{k}\IB),\\
\end{equation}
where $\traj^{k}$ denotes the motion trajectory at the $k^{th}$ noising step, and a fixed $\beta^{k}$ is chosen such that $q(\traj^{K})  \approx \mathcal{N}(\traj^{K}; \zeroB, \IB)$ after $K$ steps.
Our generative model learns the reverse of this process (denoising), \ie it recovers $\traj^{k-1}$ from a noisy input trajectory $\traj^{k}$ at each step and doing this repeatedly results in a final clean motion $\traj^{0}$. 
Because the model is generating \textit{interaction} motions with an object, we condition denoising on interaction information $C = \{ P_{o} , R_{o}, \mathbf{b}, X_0 \}$, which includes the canonicalized object point cloud $P_{o} \in \sR^{5000\times 3}$, rigid object pose relative to the person $R_{o} \in \sR^{4\times4}$, SMPL body shape parameters $\mathbf{b} \in \mathbb{R}^{10}$, and starting pose of the person $X_0$. 
Each reverse step is then:
\begin{equation}
     p_{\theta}(\traj^{k-1}| \traj^{k}, C) := \normalN(\traj^{k-1}; \mu=\mu_{\theta}(\traj^{k},k, C), \sigma=\beta^{k}\IB),
\end{equation}
where the diffusion step $k$ is also given as input. 
Instead of predicting the noise $\boldsymbol{\eps}^{k}$ added at each step of the diffusion process~\cite{ho2020denoising, janner2022diffuser}, our model directly predicts the final clean signal~\cite{tevet2022human, rempe2023trace}. 
Mathematically, the motion model $M_\theta$ with parameters $\theta$ predicts a clean trajectory $\hat{\traj}^{0}=M_{\theta}(\traj^{k}, k, C)$ from which the mean $\mu_{\theta}(\tau^{k}, k, C)$ is easily computed~\cite{nichol2021improved}. 
This formulation has the benefit that physically grounded objectives can be easily computed on $\hat{\traj}^{0}$ in the pose space, which is useful for guidance as discussed below. 

While training the diffusion model, a ground truth clean trajectory $\traj^{0}$ is noised and given as input, then the model is trained to minimize the objective $\norm{\hat{\traj}^{0} - \traj^{0}}^{2}_{2}$. To enable using classifier-free guidance~\cite{ho2022classifier} at sampling time, the conditioning $C$ is randomly masked out with 10\% probability during the training process so that the model can operate in both conditional and unconditional modes.

\parnobf{Sampling and Guidance} 
At test time, samples are generated from the model given random noise and interaction conditioning $C$ as input. 
We find that leveraging classifier-free guidance~\cite{ho2022classifier} tends to generate higher-quality samples. 
This amounts to generating one conditional and one unconditional sample from the model and then combining them as $\hat{\traj}^{0} = M_{\theta}(\traj^{k}, k) + s(M_{\theta}(\traj^{k}, k, C) - M_{\theta}(\traj^{k}, k))$, where the strength of the conditioning is controlled by the scalar $s$. 

Ensuring that the sampled motions adhere to the geometric and semantic constraints of the object is key to plausible interactions. 
Diffusion models are well-suited for this, since \textit{guidance} can encourage samples to meet desired objectives at test time~\cite{janner2022diffuser}. 
The core of guidance is a differentiable function $G(\traj^{0})$ that evaluates how well a trajectory meets a desired objective; this could be a learned~\cite{janner2022diffuser} or an analytic~\cite{rempe2023trace} function. 
In our case, we want $G(\traj^{0})$ to evaluate how plausible an interaction motion is w.r.t. the object, and in \S\ref{subsec:oif} we show that this can be done with a learned object interaction field. 
Throughout denoising during sampling, the gradient of the objective function will be used to nudge trajectory samples in the correct direction.
We use a formulation of guidance that perturbs the clean trajectory output from the model $\hat{\traj}^{0}$ at every denoising step $k$ as follows~\cite{rempe2023trace, ho2022video}:
\begin{equation}
    \label{eqn:guidance}
    \tilde{\traj}^{0} = \hat{\traj}^{0} - \alpha \nabla_{\traj^{k}}G(\hat{\traj}^{0})
\end{equation}
where $\alpha$ controls the guidance strength. 
The updated trajectory $\tilde{\traj}^{0}$ is then used to compute $\mu$. 
\parnobf{Architecture} 
As shown in \cref{fig:method} (right), the motion model $M_\theta$ is based on a transformer encoder-only architecture~\cite{vaswani2017attention,tevet2022human}. 
The model takes as input the current noisy trajectory $\traj^{k}$, the denoising step $k$, and the conditioning $C$. 
Each human pose in the trajectory is a token, while each conditioning becomes a separate token.
Of note, the object point cloud $P_{o}$ is encoded with a PointNet ~\cite{qi2017pointnet}, the rigid pose $R_{o}$ is encoded with a three-layer MLP, and $k$ is encoded using a positional embedding~\cite{tancik2020fourier}.
Our noise levels $k$ vary between 0 to 1000 diffusion steps.
The transformer handles variable-length sequence inputs and outputs the clean motion prediction $\hat{\traj}^{0}$.
Full details are available in the supplementary material.
\vspace{-2mm}
\subsection{Object Interaction Fields}
\label{subsec:oif}
After training on human-object interactions, the diffusion model can generate reasonable motion sequences but fails to fully comply with constraints in the last mile of interaction~\cite{donti2021dc3,balestriero2023police}, even when conditioned on the object. 
This causes undesirable artifacts such as penetration with the object.
\begin{wrapfigure}{r}{0.5\textwidth}
    \vspace{-4mm}
    \centering
    \begin{center}
    \includegraphics[width=0.48\textwidth]{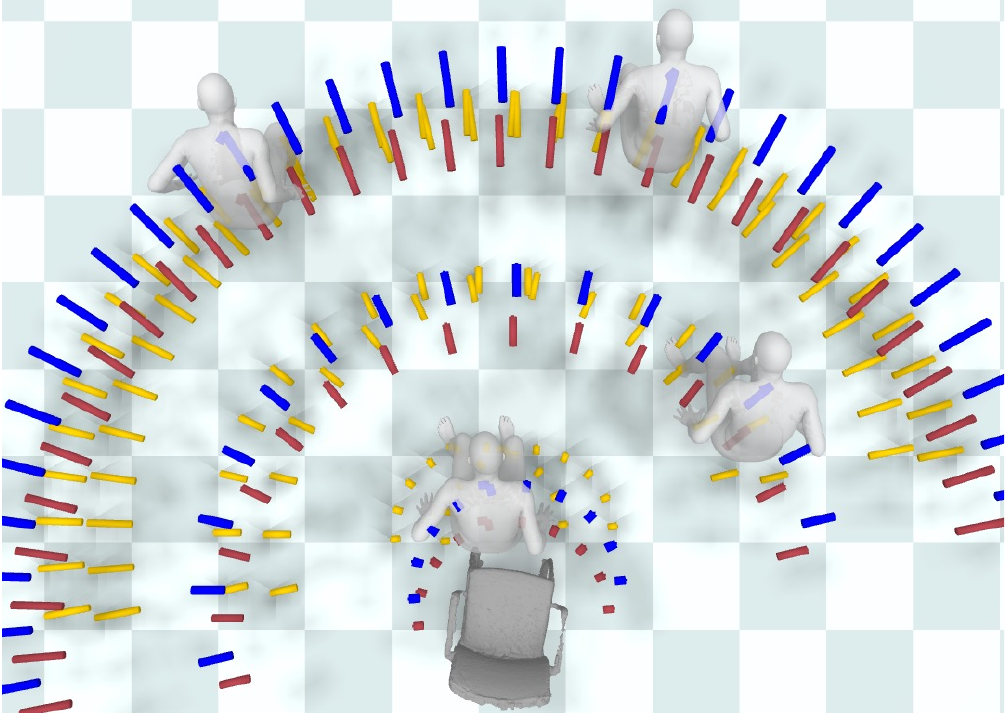} 
    \end{center}
    \vspace{-4mm}
    \caption{\textbf{Interaction Field Visualization.} We query the field in several locations with a sitting pose (a subset shown in \textcolor{grey}{grey}) and visualize the output for {\bf \textcolor{ibm_red80}{pelvis}}, {\bf \textcolor{orange}{feet}}, and {\bf \textcolor{blue}{neck}} joints. All cylinders are oriented towards the chair, indicating the correction vector's magnitude and direction. This correction is due to the misalignment between the sitting pose and chair position.
    }
    \vspace{-6mm}
    \label{fig:interaction_field}
\end{wrapfigure}
To alleviate this issue, we propose to guide motion samples from the diffusion model (\ie use \cref{eqn:guidance}) with a learned objective $G$ that captures realistic interactions for a specific object.

We take inspiration from recent work that uses neural distance fields to learn valid human pose manifolds~\cite{tiwari2022pose} and robotic grasping manifolds~\cite{weng2023ngdf}. 
For our purposes, the field must take in an arbitrary human pose and output how far the query pose is from being a ``valid'' object interaction pose. 
We define an \emph{interaction pose} to be an \emph{anchor} frame in a motion sequence that captures the core of the interaction, \eg, the moment a person settles in a chair during sitting (as in \cref{fig:overview}) or contacts an object before lifting. 

We propose an \textit{object interaction field} that operates in the local coordinate frame of a specific object.
The interaction field $F_{\phi}$ takes as input a simplified pose $\tilde{X} {=} \{{\vj^{p}, \vt^{p}}\}$, 
which includes joint positions and global translation.
The field outputs an offset \textit{vector} $\Delta \tilde{X} {=} F_{\phi}(\tilde{X})$ 
that projects the input pose to the manifold of valid interaction poses for the object: 
$\tilde{X} {+} \Delta \tilde{X}$ is then a plausible interaction pose. 
\cref{fig:interaction_field} visualizes the output vectors of an example interaction field for a chair. 
Querying  the field with a sitting pose away from the chair (\ie not a valid interaction) gives a correction pointing back towards the chair. 
For further away points, the visualized vectors are longer, indicating larger corrections are needed.

\parnobf{Guidance Objective} 
The object interaction field serves as a differentiable function that can be incorporated into the guidance objective to judge how far a motion is from the desired interaction manifold. 
Let $\tilde{X}_i \in \traj$ be the simplified pose from the $i^\text{th}$ frame of a motion $\traj$.
If we know that this pose \textit{should} be a valid interaction pose, then the guidance objective is defined as $G(\traj) = \norm{F_{\phi}(\tilde{X}_i)}_{2}^{2}$. 
During denoising at test time, we feed output poses from the diffusion model into this guidance objective to encourage the generated motion to contain a valid interaction poses. 

\parnobf{Training} 
Supervising $F_{\phi}$ requires a dataset of invalid poses with corresponding valid interaction poses. 
We collect this \textit{after} training the diffusion model detailed in \S\ref{subsec:diffusion}. 
In particular, we feed a noisy ground-truth interaction motion $\traj^k$ at a random noise level $k$ to the diffusion model as input. 
This gives an output motion $\hat{\traj}^0$, which \textit{should} match the ground truth $\traj^0$ if the model is perfect. 
In practice, denoising back to ground truth is difficult at high noise levels (\eg, $k{=}900$), so we consider $\hat{\traj}^0$ as an invalid interaction motion with a corresponding valid motion $\traj^0$.
When the diffusion model has been trained on the dataset described in \S\ref{subsec:data_gen}, we know that the last frame of the motion $\tilde{X}_N \in \hat{\traj}^0$  should be the \textit{interaction pose}, so we can throw away all other poses to arrive at a training dataset for the interaction field. 
We further augment this dataset by applying random rigid transformations to the invalid interaction poses. 

Given a ground truth interaction pose $\tilde{Y}_N \in \traj^0$ and corresponding output pose from the diffusion model $\tilde{X}_N \in \hat{\traj}^0$, 
the interaction field training loss is computed as $\norm{F_{\phi}(\tilde{X}_N)  - (\tilde{Y}_N - \tilde{X}_N)}_{1}$. 
Note that training on outputs from the diffusion model is important since the interaction field operates on these kinds of outputs during test-time guidance.

\parnobf{Architecture} 
As shown in \cref{fig:method} (left), the interaction field architecture is an encoder-only transformer that operates on the input pose as a token. 
In practice, it also takes in the canonical object point cloud as a conditioning token to allow training a single field for multiple objects.

\vspace{-2mm}
\subsection{Automatic Synthetic Data Generation}
\begin{figure}[t]
    \centering
    \noindent
  
    \includegraphics[width=0.95\textwidth]{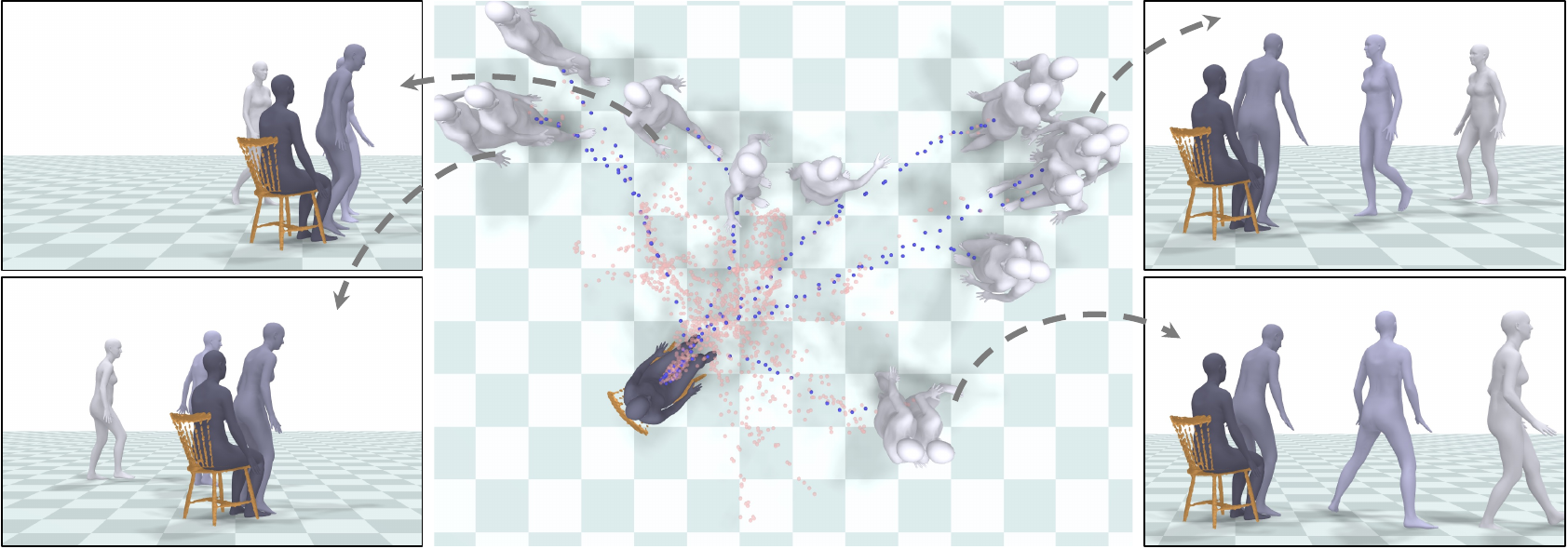}
    \vspace{-2mm}
    \caption{{\bf Generated Synthetic Data}. We visualize motion sequences from one tree rollout for one sitting {\bf \textcolor{lastblue}{ anchor pose}}. The middle shows a bird's-eye view of the {\it pelvis} joint trajectories in {\bf \textcolor{pink}{light pink}}. All trajectories end in the same sitting pose, but start at diverse locations around the chair. We highlight a few trajectories in {\bf \textcolor{blue}{blue}} and show full-body motions from the corresponding generations on the left and right sides. Our complete dataset contains many trees for different objects and humans. 
    }
    \label{fig:diversity}
    \vspace{-5mm}
\end{figure}
\label{subsec:data_gen}
Training the diffusion model requires a large, realistic, and diverse dataset of motions for each human-object interaction we wish to synthesize.
Unfortunately, this data exists only for specific interactions~\cite{zhang2022couch} and is difficult and expensive to collect from scratch.
Therefore, we propose an automated pipeline to generate synthetic interaction motion data. 
In short, we first select \textit{anchor} pose frames from an existing small dataset~\cite{bhatnagar22behave} that are indicative of an interaction we want to learn. 
Our key insight is to use a pre-trained scene-unaware motion model~\cite{rempe2021humor} to sample a diverse set of motions that \textit{end} at a selected anchor pose, and therefore demonstrate the desired interaction. We provide the key details in this section and a full description appears in the supplementary material.

\parnobf{Anchor Pose Selection}
We require a small set of anchor poses that capture the core frame of an interaction motion. 
As described in \S\ref{subsec:oif}, for sitting on a chair this is the sitting pose when the person first becomes settled in the chair (see \cref{fig:overview}). 
In generating motion data, these anchor poses will be the \textit{final} frame of each synthesized motion sequence. 
For the experiments in \S\ref{sec:experiments}, these anchor frames are chosen manually from a small dataset that contains a variety of interactions~\cite{bhatnagar22behave}. 

\parnobf{Generating Motions in Reverse} 
The goal is to generate human motions that \textit{end} in the chosen anchor poses and reflect realistic object interactions.
We leverage HuMoR~\cite{rempe2021humor}, which is a conditional motion VAE trained on the AMASS~\cite{mahmood2019amass} mocap dataset. It generates realistic human motions through autoregressive rollout, but is \textit{scene-unaware}.
To force rollouts from HuMoR to match the final anchor pose, we could use online sampling or latent motion optimization, but these are expensive and not guaranteed to exactly converge. 
Instead, we re-train HuMoR as a time-reversed motion model that predicts the past instead of the future motion given a current input pose. 
Starting from a desired interaction anchor pose $X_N$, our reverse HuMoR will generate $X_{N-1}, X_{N-2}, \cdots, X_{1}$ forming a full interaction motion that, by construction, ends in the desired pose. 

\parnobf{Tree-Based Rollout \& Filtering} 
To ensure sufficient diversity and realism in motions from HuMoR, we devise a branching rollout strategy that is amenable to frequent filtering and results in a tree of plausible interactions. 
Starting from the anchor pose, we first sample $30$ frames (1 sec) of motion. 
Then, multiple branches are instantiated and random rollout continues for another $30$ frames on these branches independently. 
Continuing in this branching fashion allows growing the motion dataset exponentially while also filtering to ensure branches are sufficiently diverse and do not contain undesireable motions. 
Filtering involves heuristically pruning branches with motions that collide with the object, float above the ground plane, result in unnatural pose configurations, and become stationary. 
For the experiments in \S \ref{sec:experiments}, we rollout to a tree depth of 7 and sample many motion trees starting from each anchor pose. 
Individual paths are extracted from the tree to give interaction motions, and we post-filter out sequences that start within 1 meter of the object.

\parnobf{Generated Datasets}
We use this scalable strategy to generate data for training our motion model for \textit{sitting} and \textit{lifting} interactions. 
\cref{fig:diversity} demonstrates the diversity of our generated datasets by visualizing top-down trajectories and example motions from a single tree of sitting motions.
For the sitting interaction dataset, we choose 174 anchor pose frames across 7 subjects in the BEHAVE~\cite{bhatnagar22behave} dataset. 
This results in a dataset of 200K motion sequences that include sitting on chairs, stools, tables, and exercise balls. Each motion sequence in this dataset ends at a sitting anchor pose.
For lifting interactions, 72 anchor poses from 7 subjects produces 110K motion sequences. Each sequence ends at a lifting anchor pose when the person initially contacts the object.

\vspace{-2mm}
\section{Experiments}

\label{sec:experiments}

We evaluate our \ours method after training on the {\it sitting} and {\it lifting} datasets introduced in \S\ref{subsec:data_gen}. 
Implementation details are given in \S\ref{subsec:impl}, followed by a discussion of evaluation metrics in \S\ref{subsec:eval} and baselines in \S\ref{subsec:baselines}. 
Experimental results are presented in \S\ref{subsec:results} along with an ablation study in \S\ref{subsec:ablation}.

\vspace{-1mm}
\subsection{Implementation Details}
\label{subsec:impl}
\vspace{-1mm}
We train our diffusion model $M_\theta$ for 600K iterations with a batch size of 32 using the AdamW~\cite{loshchilov2017decoupled} optimizer with a learning rate of $10^{-4}$. A separate model is trained for sitting and lifting. 
We use $K{=}1000$ diffusion steps in our model and sample the diffusion step $k$ from a uniform distribution at each training iteration.
The object interaction field $F_{\phi}$ is trained on the data described in \S\ref{subsec:oif} for 300K iterations using AdamW with a maximum learning rate of $5\times10^{-5}$ and a one cycle LR schedule~\cite{smith2019super}. 
When sampling from the diffusion model, 10 samples are generated in parallel and all are guided using the object interaction field; the sample with the best guidance objective score is used as the output. 
We apply interaction field guidance on the last frame of motion (\ie the interaction anchor pose in our datasets). 
Our models are trained using PyTorch~\cite{paszke2019pytorch} on NVIDIA A40 GPUs, and take about 2 days to train. Visualizations use the PyRender engine~\cite{pyrender}.

\vspace{-2mm}
\subsection{Evaluation Setting and Metrics}
\label{subsec:eval} 
To ensure we properly evaluate the generalization capability of methods trained on our synthetic interaction datasets, we \textit{do not} create a test set using the procedure described in \cref{subsec:data_gen}, which may result in a very similar distribution to training data. 
Instead, we create a set of 500 \textit{test scenes} for each action, where objects are randomly placed in the scene and the human starts from a random pose generated by HuMoR. 
All methods are tested on these same scenes during evaluation.

Evaluating human motion coupled with object interactions is challenging and has no standardized protocol. 
 Hence, we evaluate using a diverse set of metrics including a user perceptual study. 
 We briefly describe the metrics next and include full details in the supplementary material.

 \parnobf{User Study} 
 No single metric can capture all the nuances of human-object interactions, so we employ a perceptual study~\cite{mir2023generating, tevet2022human, taheri2022goal, taheri2020grab, wang2022humanise}. 
For each method, we create videos from generated motions on the test scenes.
 To compare two methods, users are presented with two videos on the same test scene and must choose which they prefer (full user directions are in the supplement). 
 We perform independent user studies for lifting and sitting actions using \texttt{hive.ai}~\cite{hive}. 
 Responses are collected from 5 users for every comparison video, giving 2500 total responses in each comparison study.
 
 \parnobf{Foot Skating} 
 Similar to prior work~\cite{mir2023generating}, we define the foot-skating score for a sequence of $N$ timesteps as $\frac{1}{N}\sum_{i}^{N}v_{i}(2 - 2^{h_{i}/H}) \cdot \mathbbm{1}_{h<=H}$, where $v_{i}$ is the velocity and $h_{i}$ is the height above ground of the right toe vertex for the $i^{th}$ frame. $H$ is $2.5$ cm. 
 Intuitively, this is the mean foot velocity when it is near the ground (where it \emph{should} be 0), with higher weight applied closer to the ground. 
 
 \parnobf{Distance to Object ({\it D2O})} 
 Similar to prior work~\cite{wang2022humanise}, this evaluates whether the human gets close to the object during the interaction.  
 It measures the minimum distance from the human body in the last frame of the motion sequence to any point on the object's surface. 
  We report the \% of sequences within $2$ cm distance to avoid sensitivity to outliers, along with the $95^\text{th}$ percentile ($\tilde{\%}$) of this distance. 

 \parnobf{Penetration Score ({\it \% Pen)}} 
 To evaluate realism as the human approaches an object for interaction, we measure how much they penetrate the object.
 Based on our synthetic data, we define the first $N_{A}$ frames of motion to be the approach for each action type (see supplement). 
 
Then the penetration distance for a trajectory is $\frac{1}{N_{A}}\sum_{v}\sum_{i}^{N_{A}}\texttt{sdf}_{i}(v) \cdot \mathbbm{1}_{\texttt{sdf}_{i}(v) > 0}$, where $\texttt{sdf}_{i}$ is the signed distance function of the human in the $i^{th}$ frame and $v$ is one of 2K points on the object's surface. 
 We report the percentage of trajectories with penetration distance $\le$ 2 cm (\% Pen. $\le 2cm$) ignoring trajectories with {\it D2O} $>$ 2 cm, since trajectories that do not approach the object will trivially avoid penetration. 
 
\parnobf{Skeleton Distance \& Contact IoU} 
These evaluate how well generated interaction poses align with ground truth poses and their human-object contacts. 
We start by finding the minimum distance between the final pose of a generated sequence and the anchor poses in the synthetic training data. 
The distance to this nearest neighbor pose is reported as the skeleton distance. 
To measure how well contacts from the generated motion match the data, we compute the IoU between contacting vertices (those that penetrate the object) on the predicted body mesh and those on the nearest neighbor mesh.

\begin{table}[t]
    \setlength{\tabcolsep}{3pt}
    \centering 
     \caption{{\bf Quantitative Comparison.} Our method outperforms baselines on both sitting and lifting. Our diffusion model, guided by the learned interaction field, generates motions that reach the object (\textit{D2O}) with few \textit{penetrations} and realistic \textit{contacts}. Motions approaching the object are realistic with low \textit{foot skating} and the final interaction pose is similar to synthetic data with low \textit{skeleton distance}.} 
     \vspace{-1mm}
    \begin{adjustbox}{max width=\linewidth}
     \begin{tabular}{l r@{~~}r@{~~}r@{~~}rrr|@{~}s@{~~}s@{~~}s@{~~}sss} \toprule
        
                & \multicolumn{6}{r}{\textbf{Sitting}} & \multicolumn{6}{s}{\textbf{Lifting}} \\
               
        Method  & Foot                 & \% {\it D2O}          &  {\it D2O}                             & Skel.                & Contact               & \% Pen.              & Foot               & \%  {\it D2O}                &  {\it D2O}                           & Skel.               & Contact             & \% Pen.    \\
                & Skating $\downarrow$ & $\le 2cm$ $\uparrow$  &  $95^\text{th}\tilde{\%}$ $\downarrow$  & Dist. $\downarrow$   & IoU    $\uparrow$  &  $\le 2cm$ $\uparrow$   & Skating $\downarrow$  & $\le 2cm$ $\uparrow$  &   $95^\text{th}\tilde{\%}$ $\downarrow$  & Dist. $\downarrow$   & IoU    $\uparrow$  &  $\le 2cm$ $\uparrow$ \\
                 \midrule
        \humanise & 0.77 & 88.8 & 0.13 & 1.07 & 0.21 & 46.6     &       0.66 & 57.4 & 0.66 & 1.70 & 0.03 & 60.1 \\
        \mdmOnly  & \textbf{0.36} & 37.9 & 1.06 & 2.81 & 0.04 & 50.5     &       \textbf{0.28} & 36.3 & 1.14 & 2.58 & 0.02 & 46.2 \\
         \ours (ours)    & 0.47 & \textbf{99.6} & \textbf{0.00} & \textbf{0.54} & \textbf{0.54} & \textbf{65.0}    &       0.34 & \textbf{77.7} & \textbf{0.05} & \textbf{0.42} & \textbf{0.17} & \textbf{68.5} \\
        \midrule
    \end{tabular}
    \end{adjustbox}
   
    \label{tab:quant}
    \vspace{-7mm}
\end{table}

\vspace{-2mm}
\subsection{Baselines}
\label{subsec:baselines}

\parnobf{Cond. VAE \cite{wang2022humanise}} 
Closest to our problem definition, this model comes from recent work HUMANISE~\cite{wang2022humanise}, which learns plausible human motions conditioned on scene and language for four actions (lie, sit, stand, walk). 
This state-of-the-art model is a conditional VAE with a GRU motion encoder and sequence-level transformer decoder.
Since we evaluate on sitting and lifting actions separately, we modify their approach to remove language conditioning. 
The model is trained on our synthetic data for 600K iterations with the recommended hyperparameters and learning rate of $10^{-4}$.

\parnobf{\mdmOnly} 
This baseline is the motion diffusion model (MDM)~\cite{tevet2022human} with added conditioning $C$, \ie our object-conditioned diffusion model without interaction field guidance.  
\begin{wrapfigure}{r}{0.45\textwidth}
    \vspace{-5mm}
    \centering
    \begin{center}
    \includegraphics[width=0.40\textwidth]{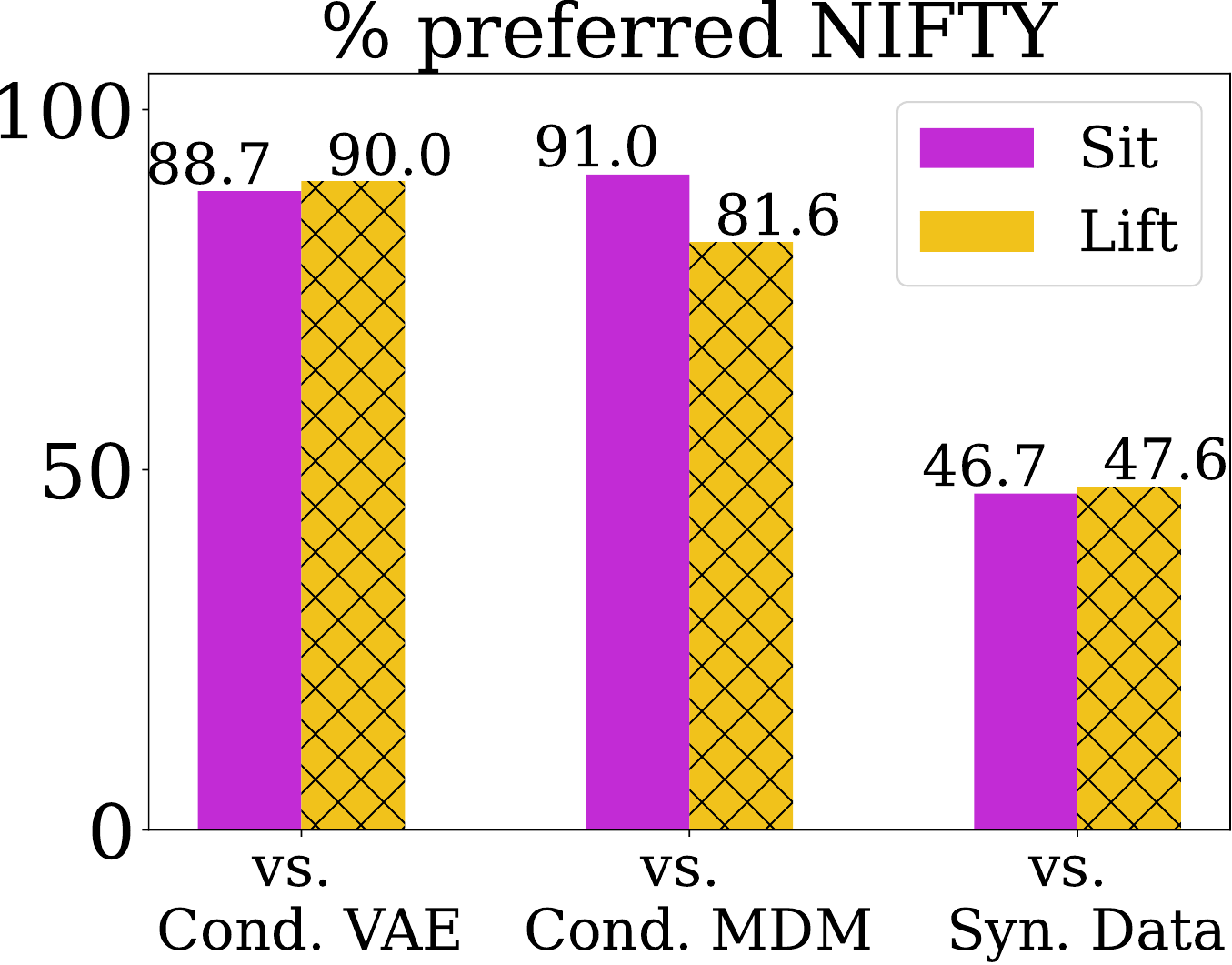} 
    \end{center}
    \vspace{-4mm}
    \caption{\textbf{User Study.} 
    \ours is preferred $\ge$ 88.7\% of the time for sitting and $\ge$81.6\% for lifting compared to baselines. Our motions are also nearly indistinguishable from synthetic data trajectories. 
    }
    \vspace{-9mm}
    \label{fig:user_study}
\end{wrapfigure}

\vspace{-4mm}
\subsection{Experimental Results}
\label{subsec:results}

\parnobf{User Study} 
\cref{fig:user_study} shows how often users prefer our method (\ours) over baselines and Synthetic Data (Syn. Data) for both sitting and lifting. 
We perform separate studies for each comparison. Users prefer \ours over baselines a vast majority of the time. Averaged 
over both actions, \ours is preferred over the \humanise baseline 89.4\% of the time. 
Similarly, \ours is preferred over \mdmOnly 86.3\% of the time, highlighting the importance of using guidance with our interaction field during sampling. 
Compared to held out motions from synthetic data, \ours is preferred $47.2\%$ of the time, which indicates that the motions are nearly indistinguishable from those of our data generation pipeline. 

We further extract robust consensus across users by majority vote over the 5 responses for each video. 
In this case, motions generated by our method are preferred 94.4\% (sitting) and 97.8\% (lifting) of the time over \humanise motions, making the improvement gap even more apparent.

Additionally, we also conduct an user study on a Likert scale of scores between 1 (unrealistic) to 5 (very realistic). We report that motions from our synthetic dataset achieve a score of 4.39 {\it vs.} 4.87 for motions in the AMASS~\cite{mahmood2019amass}. Further details are available in Supp. \S~\ref{subsec:data_quality}.

\parnobf{Quantiative Results} 
In \cref{tab:quant}, \ours is compared to baselines for both sitting and lifting interactions.
\ours generates motions that reach the target object and approach realistically, as indicated by distance-to-object (\textit{D2O}) and \textit{penetration} metrics. 
Although \mdmOnly produces realistic motion with low \textit{foot skating}, it struggles to properly approach the object since it does not use guidance from the learned interaction field. 
We see that interaction poses and the resulting object contacts generated by our method do reflect the synthetic dataset, resulting in low \textit{skeleton distance} and high \textit{contact IoU}, unlike \humanise which is worse across all metrics. 

\begin{figure}[t]
    \centering
    \noindent
    \begin{adjustbox}{max width=\linewidth}
    \begin{tabular}{c}
    \noindent\includegraphics{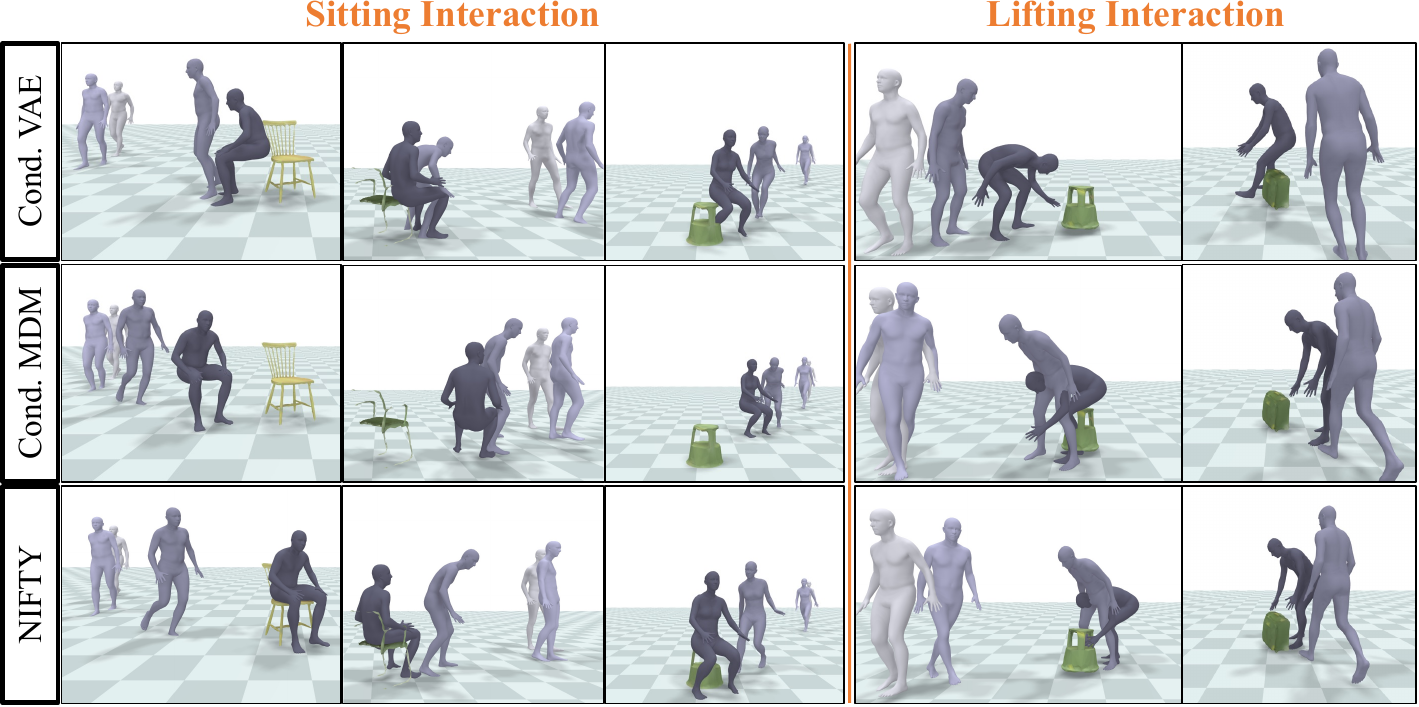} 
    \end{tabular}
    \end{adjustbox}
    \vspace{-2mm}
    \caption{\textbf{Qualitative Results.} Our method (bottom row) generates realistic interaction motions that reach the desired object with plausible contacts (\eg{~}col 1 \& 4) while avoiding penetrations, unlike baselines. The mesh color gets darker as time progresses. \humanise motions have the {\bf \textcolor{lastblue}{final interaction pose}}  away from the object (col 1,3,4), incorrect (col 2 \& 5), or intersecting (col 5). \mdmOnly generates sitting poses far away from the object (col 1 \& 3). 
    }
    \label{fig:qual}
    \vspace{-5mm}
\end{figure}

\parnobf{Qualitative Results}
\cref{fig:qual} shows a qualitative comparison between motions generated by our method and baselines. 
\ours synthesizes realistic sitting and lifting with a variety of objects. 
Examples show that the baselines struggle to generalize to unseen object poses, and have no mechanism to correct for this at test time. 
Our learned interactions field helps to avoid this through diffusion guidance. Please see the videos provided in the supplement to best appreciate the results.

\begin{table}[t]
    \setlength{\tabcolsep}{3pt}
    \caption{\textbf{Ablation Study.} Comparison between using an interaction field trained to predict a full offset vector (\ours) or a single scalar distance ({Distance OIF}).} 
    \centering 
    \vspace{-1mm}
    \begin{adjustbox}{max width=\linewidth}
    
     \begin{tabular}{l r@{~~}r@{~~}r@{~~}rrr|@{~}s@{~~}s@{~~}s@{~~}sss} \toprule
        
                & \multicolumn{6}{r}{\textbf{Sitting}} & \multicolumn{6}{s}{\textbf{Lifting}} \\
        Method  & Foot                 & \% {\it D2O}          &  {\it D2O}                    & Skel.                & Contact               & \% Pen.              & Foot               & \%  {\it D2O}            &  {\it D2O}                    & Skel.               & Contact             & \% Pen.    \\
                & Skating $\downarrow$ & $\le 2cm$ $\uparrow$  &  $95^\text{th}\tilde{\%}$ $\downarrow$  & Dist. $\downarrow$   & IoU    $\uparrow$  &  $\le 2cm$ $\uparrow$   & Skating $\downarrow$  & $\le 2cm$ $\uparrow$  &   $95^\text{th}\tilde{\%}$ $\downarrow$  & Dist. $\downarrow$   & IoU    $\uparrow$  &  $\le 2cm$ $\uparrow$ \\
                 \midrule
        Distance OIF & \textbf{0.41} & 80.9 & 0.47 & 1.25 & 0.24 & \textbf{66.8}               & \textbf{0.30} & 57.4 & 0.74 & 1.31 & 0.07 & \textbf{70.1}\\
         \ours  & 0.47 & \textbf{99.6} & \textbf{0.00} & \textbf{0.54} & \textbf{0.54} & 65.0         & 0.34 & \textbf{77.7} & \textbf{0.05} & \textbf{0.42} & \textbf{0.17} & 68.5 \\
        \midrule
    \end{tabular}
    \end{adjustbox}
  
    \label{tab:abl}
    \vspace{-3mm}
\end{table}

\vspace{-2mm}
\subsection{Ablation Study}
\label{subsec:ablation} 
As detailed in \S\ref{subsec:oif}, our object interaction field (OIF) is formulated to predict an offset vector $\Delta\tilde{X}$ that captures both distance and direction for each component of the pose state, rather than a single full-body distance like prior work~\cite{tiwari2022pose}. 
We ablate this design decision in \cref{tab:abl}, which compares our interaction field formulation to a version that predicts only a scalar distance to the interaction pose manifold ({Distance OIF}). 
We observe that learning a single distance is a harder task compared to predicting an offset vector, which provides a stronger learning signal for training. 
As a result, the ablated interaction field results in worse scores across most metrics.

\vspace{-2mm}
\subsection*{5. Conclusion and Limitations}
\vspace{-2mm}
\label{sec:conclusion} 
We introduced \ours, a framework for learning to synthesize realistic human motions involving 3D object interactions. 
Results demonstrate that our object-conditioned diffusion model gives improved motions over prior work when guided by a learned object interaction field and trained on automatically synthesized motion data. 
Our current approach is limited to the body shapes present in the training data (\eg,  the 7 subjects from BEHAVE~\cite{bhatnagar22behave}), so future work should explore data augmentation strategies to generalize to novel humans. 
Moreover, we have shown results on sitting and lifting, but we would like to widen the scope to handle additional interactions by collecting new anchor poses, synthesizing data, and training our diffusion model and interaction field.  

\parnobf{Acknowledgements}. We express our gratitude to our colleagues for the fantastic project discussions and feedback provided at different stages. We have organized them by institution (in alphabetical order) \\
-- {\it Google}: Matthew Brown, Frank Dellaert, Thomas A. Funkhouser, Varun Jampani\\
-- {\it Google (co-interns)}: Songyou Peng, Mikaela Uy, Guandao Yang, Xiaoshuai Zhang \\
-- {\it University of Michigan}:  Mohamed El Banani, Ang Cao,  Karan Desai,  Richard Higgins,  Sarah Jabbour, Linyi Jin, Jeongsoo Park, Chris Rockwell, Dandan Shan

This work was partly done when NK was interning at Google Research. DR was supported by the NVIDIA Graduate Fellowship.

{\small
\bibliographystyle{plain}
\bibliography{egbib}
}
\clearpage
\appendix
\section{Automated Synthetic Training Data Generation}
\label{sec:datagen}
All models in the paper train on synthetic human-object interaction motion data generated using this pipeline. To evaluate the quality of generated data compared to other data, in \S~\ref{subsec:data_quality} we perform a large scale user-study with 10K user responses. 
In \S~\ref{subsec:algo} we describe the complete details of data generation including pseudo-code for the algorithm.

\subsection{Data Quality User Study}
\label{subsec:data_quality}
\begin{figure}[b]
    \centering
    \noindent
    \begin{adjustbox}{max width=\linewidth}
    \begin{tabular}{c}
    \noindent\includegraphics{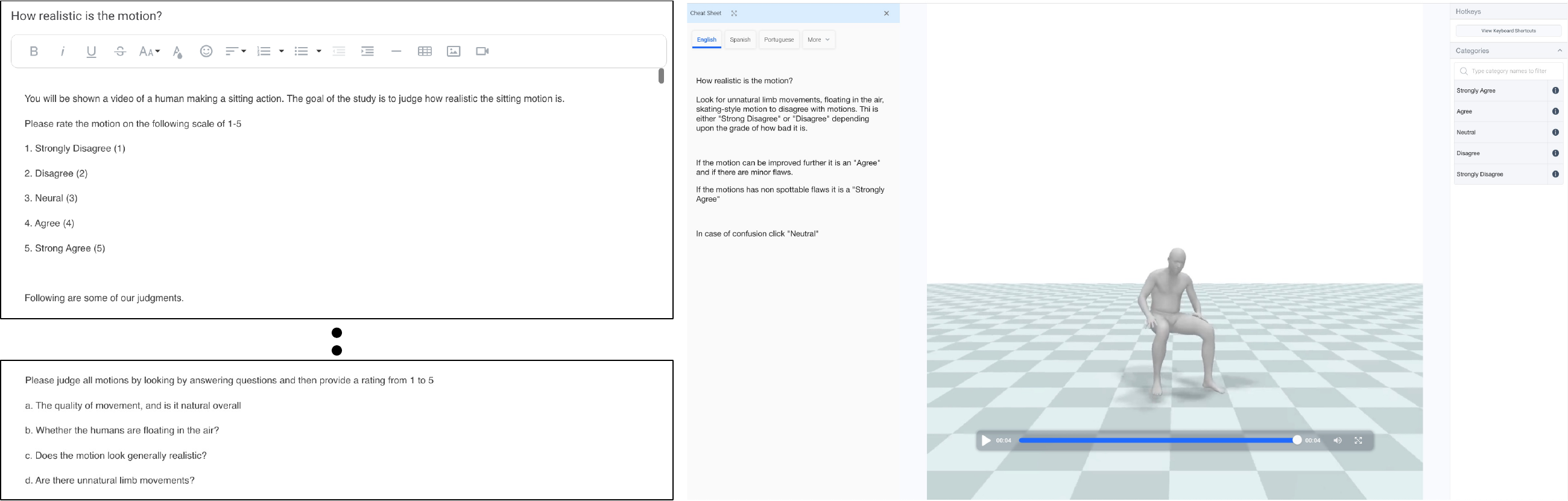} 
    \end{tabular}
    \end{adjustbox}
    \vspace{-2mm}
    \caption{\textbf{Likert User Study.} We conduct a user study to assess the motion quality in our Synthetic Dataset. On the left, we present the qualification instructions for participants, allowing only those who perform well to proceed to the actual study. On the right, we display the user interface used for labeling motions, where users select from five options: ``Strongly Agree", ``Agree", ``Neutral", ``Disagree", or ``Strongly Disagree". The results of this study can be found in \cref{fig:likert_results}
    }
    \label{fig:likert_study}
\end{figure}
Our synthetic data generation pipeline helps us collect high-quality motion data corresponding to different interaction anchor poses. We show that this generated data is high-quality by conducting a user study on a five-point Likert scale as in prior work~\cite{taheri2020grab, taheri2022goal}. Our results show that the generated synthetic training data is on par with data collected using a real mocap setup.

\parnobf{User Study Setup} 
We created a user-study dataset of $2000$ videos, consisting of $500$ motions from the AMASS subset of HUMANISE sitting data~\cite{wang2022humanise} (\ie real-world \textit{motion captured} data), $500$ motions from our data generation pipeline, $500$ predicted motions from our \ours sitting model, and $500$ from \humanise predictions. For each motion, we rendered a video without an object present in the scene to make the source of the video indistinguishable. All motions had a random number of frames uniformly sampled from $60$ to $120$, where the last motion frame always corresponded to the sitting interaction pose. We only show results on sitting as the HUMNISE~\cite{wang2022humanise} does not have lifting interaction AMASS subset in their data.

We ask the users to rate the video on its realism. Users are asked to rate on a scale of $1$ to $5$ corresponding to ``Strongly Disagree", ``Disagree", ``Neutral", ``Agree", ``Strongly Agree". We set up the study on \texttt{hive.ai}~\cite{hive}. Instructions to the user are shown in \cref{fig:likert_study}.

\begin{figure}[t]
    \centering
    \noindent
    \begin{adjustbox}{max width=0.45\linewidth}
    \begin{tabular}{c}
    \noindent\includegraphics{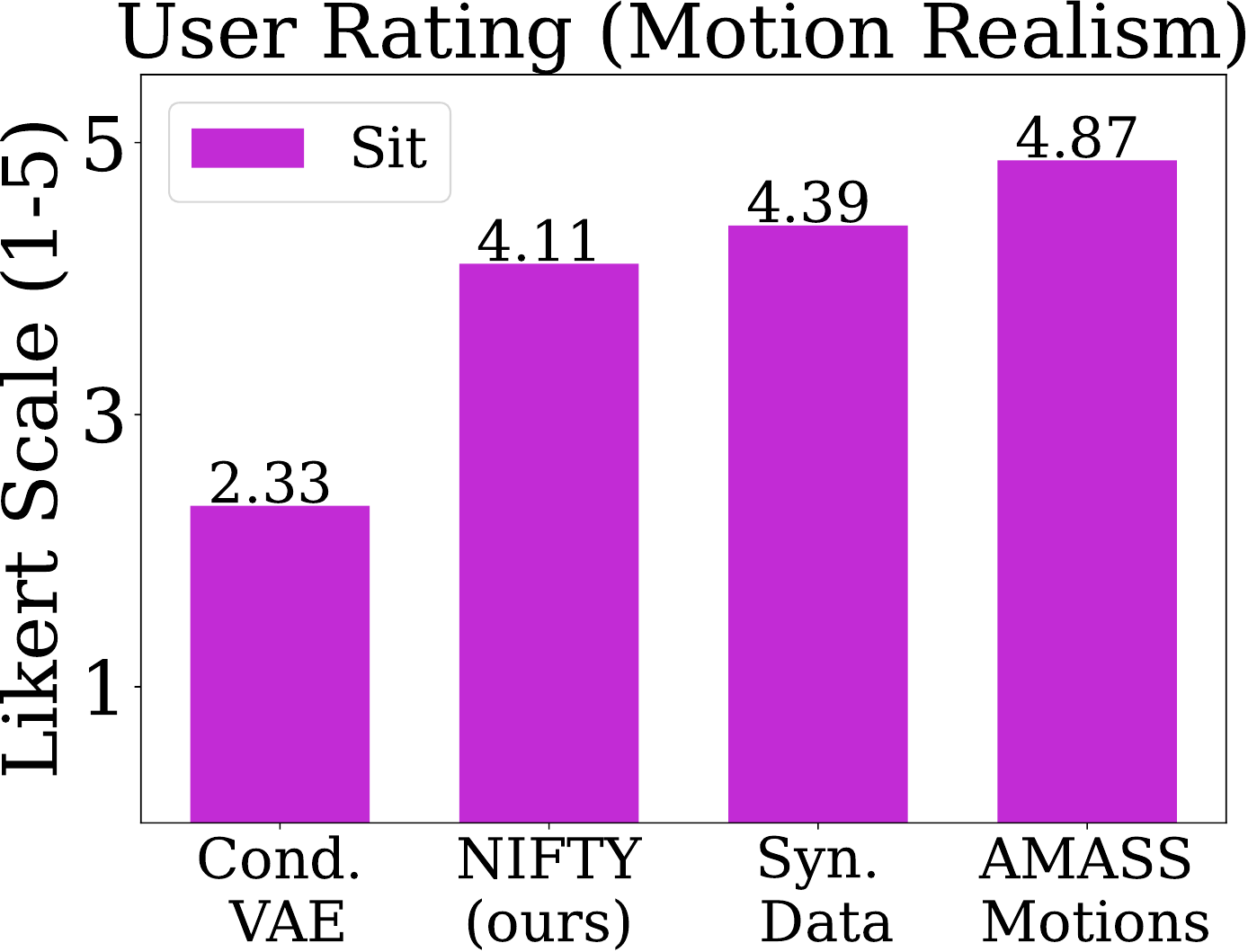} 
    \end{tabular}
    \end{adjustbox}
    \vspace{-2mm}
    \caption{\textbf{Likert User Study Results.} We conduct a study to judge the realism of sitting motions on a scale of 1-5. Instructions for this study are available in \S\ref{subsec:data_quality}. We show that synthetic training data (Syn. Data) generated using our algorithm \cref{alg:tree_generation} has an average rating of {\bf 4.39}. This is comparable to AMASS motions which represent quality of real captured data (using a mocap setup).
    }
    \label{fig:likert_results}
\end{figure}

\parnobf{User Study Results} 
 Results are shown in  \cref{fig:likert_results}.
 As expected, AMASS has a high realism score of 4.87 since it is actual mocap data. 
 Training data generated using our algorithm has an average user rating of 4.39, implying the quality is comparable motion collected using an expensive mocap setup. We also report the performance of \ours and \humanise methods on the same study for completeness. \ours achieves a strong score of 4.11 (between ``agree'' and ``strongly agree''), which is close to score of the Syn. Data. 
 The  \humanise performance remains low at 2.33 (between ``disagree'' and ``neutral'').


\parnobf{Filtering Unreliable Users} 
Note that every user is required to pass a qualification test containing easy examples to label. 
User accuracy is computed and users with accuracy > 60\% are admitted. To ensure that we collect valid responses and that users completely understand the task during the actual study, they are occasionally tested on ``obvious" data called ``honey pots" during the labeling process. To this end, we add motions with objective ``Strongly Agree" labels (motions from AMASS) and some with "Strongly Disagree" labels (low-quality motions generated by cVAE).  
This is common practice while conducting such a study, and we also do this for the user study in the main paper as detailed in \S\ref{subsec:userstudy-details}. The honeypot accuracy for this task is set at 82\%: drops in performance below this thresholds removes a user from continuing the study any further. 

\subsection{Training Data Synthesis Algorithm}
\label{subsec:algo}
Our generation process revolves around utilizing a pretrained motion model, specifically the HuMoR generative model~\cite{rempe2021humor}, to produce motion trajectories that \textit{end} in a specific anchor pose. However, we train this model on reverse-time sequences, enabling us to generate reverse-time sequences that \textit{start} from the provided anchor seed pose. Then, when we convert these rollouts into forward motions (\ie play them backwards), the final generated pose in the rollout aligns with the anchor pose by design.

Our full algorithm for generating a single motion tree is shown in \cref{alg:tree_generation}. 
This algorithm constructs a tree of a specified depth, where each node corresponds to a 1 sec motion clip. Each node is connected to several possible branches to continue the motion (based on a branching factor $B$). The algorithm begins by creating a root node starting at an input anchor pose. It then repeatedly constructs the tree by generating motion sequences using the \texttt{RollOut} function and checking their validity using the  \texttt{PruneCheck} function. If a valid motion sequence is obtained, a child node is created and added to the tree. The process continues until the desired depth is reached or the tree is fully explored (no more branches left to explore)

The algorithm maintains a queue of nodes to be processed, allowing for breadth-first construction of the tree. If a node reaches the maximum depth, it is skipped to ensure the tree is constructed as per the specified depth. The algorithm outputs the resulting tree, which contains valid motion sequences as paths from the root to the leaf nodes.

\begin{algorithm}

\caption{{\bf Tree Generation}. Our proposed tree-roll out algorithm using a pre-trained motion-model}  \label{alg:tree_generation}
\begin{algorithmic}[1]

\Function{RollOut}{$\text{startPose}$, $N$} \Comment{Input: start pose, $N$ defining number of rollout attempts}
  \State $\text{validSequence} \gets \text{False}$
  \State $\text{count} \gets 0$

  \While{$\text{not validSequence}$ \textbf{and} $\text{count} < N$}
    \State $\text{motion} \gets \text{pretrained motion model generate motion using startPose}$
    \State $\text{validSequence} \gets \text{PruneCheck}(\text{motion})$
    \State $\text{count} \gets \text{count} + 1$
  \EndWhile

  \If{$\text{validSequence}$}
    \State \Return $\text{motion}$
  \Else
    \State \Return $\text{null}$
  \EndIf
\EndFunction

\item[]

\Function{PruneCheck}{$\text{motionSequence}$} \Comment{Input: motion sequence}
  \State $\text{valid} \gets \text{check if motionSequence is valid}$
  \State \Return $\text{valid}$
\EndFunction

\item[]

\State $\text{queue} \gets \text{empty queue}$

\State $\text{rootAnchorPose} \gets \text{input anchor pose}$
\State $\text{root} \gets \text{create root node NULL motion} $ \Comment{For the root node there is no past motion (NULL).}

\State $\text{root.lastPose} \gets \text{root.anchorPose} $  \Comment{The anchor pose is the seed for future roll-outs}

\State $\text{queue.push}(\text{root})$

\While{$\text{queue is not empty}$}
  \State $\text{currentNode} \gets \text{queue.pop()}$

  \If{$\text{currentNode.depth} = \text{MaxDepth}$}
    \State \textbf{continue}
  \EndIf

  \For{$\text{child} \gets 1$ \textbf{to} $B$}
    \State $\text{GMotion} \gets \text{RollOut}(\text{currentNode.lastPose}, \text{NTries})$ \Comment{Create a RollOut}

    \If{$\text{GMotion} \neq \text{null}$} \Comment{Check if Good RollOut?}
      \State $\text{childNode} \gets \text{create child node with GMotion}$
      \State $\text{childNode.lastPose} \gets \text{GMotion last frame}$ \Comment{Set the last motion frame for childNode}
      \State $\text{currentNode.children.push}(\text{childNode})$
      \State $\text{queue.push}(\text{childNode})$  \Comment{Add childNode to queue}
    \EndIf
  \EndFor
\EndWhile

\end{algorithmic}
\end{algorithm}

\parnobf{RollOut Function} The \texttt{RollOut} function takes an start pose and utilizes the pre-trained motion model to generate a short 1 sec (30 frame) motion sequence. It iteratively runs the motion model until a valid sequence is obtained or a specified maximum number of attempts is reached. If a valid sequence is found, it is returned as the generated motion.
\parnobf{PruneCheck Function} The \texttt{PruneCheck} function examines a given motion sequence to determine its validity. It algorithmically checks if the motion collides with the object, has unnatural human poses, if the human is floating in the air, or intersecting with the floor \etc. It returns a boolean value indicating whether the motion sequence is valid or not.


\parnobf{Implementation} In our implementation, we set $B$ as $6$ for the nodes at depths $1$ and $2$, while $B=2$ for nodes at higher depths. We also set \texttt{NTries} as $20$ to secure a good rollout sequence. We then convert all the motion nodes in these trees into individual motion sequences for a particular interaction.
\section{Implementation Details}
\label{sec:implementation}

\parnobf{Recovering Motion from $\traj^{0}$} Our trajectory representation is over-parameterized and this allows using the model outputs in multiple ways. To recover the generated motion we extract the per-frame joint angles $\vj^{r}_{i}$ for the SMPL model. We integrate the velocity $\vt_{i}^{v}$ along the XZ plane to recover the XZ translation for the root joint and extract the corresponding Y component (upward) from  $\vt_{i}^{p}$.  
This strategy of extracting motion from the output parameterization is motivated by our use of guidance with the diffusion model, which only operates on the last frame of a motion sequence. By integrating velocity predictions over time, applying the guidance objective at the last frame will still have a strong effect on earlier frames in the sequence.

\parnobf{Variable Length Input} Our model takes input motion trajectories with up to 150 frames. For training, we pad motion sequences of lengths shorter than this with the last interaction frame from the sequence. 

\parnobf{SMPL model} Our SMPL~\cite{smpl} model does not have hand articulation, so we use the SMPL model with only $22$ articulated joints.

\parnobf{Pre-trained Motion Model for Data Generation} 
We train the motion model on a subset of the AMASS dataset that does not contain extreme sporting actions like jumping, dancing, \etc We do this by removing sequences from AMASS based on the labels from the BABEL dataset~\cite{punnakkal2021babel}. We use the HuMoR-Qual~\cite{rempe2021humor} variant of the model to get high-quality motions, which uses the joint positions computed through the SMPL parametric model as input to future roll-out time steps (as opposed to using its own joint position predictions).

\parnobf{Transfomer Encoder}. We use a transformer encoder implemented using \texttt{torch.nn.TransformerEncoder} from PyTorch~\cite{paszke2019pytorch}. Our  each transformer layer consists of 4 heads and a latent dim on $512$. We have 8 such layers in our transformer.
\section{Experimental Details}
\label{sec:quantitative}
This section provides additional details on the implementation of our user study and metrics from the main paper in \S 4. 

\begin{figure}[h!]
    \centering
    \noindent
    \begin{adjustbox}{max width=\linewidth}
    \begin{tabular}{c}
    \noindent\includegraphics{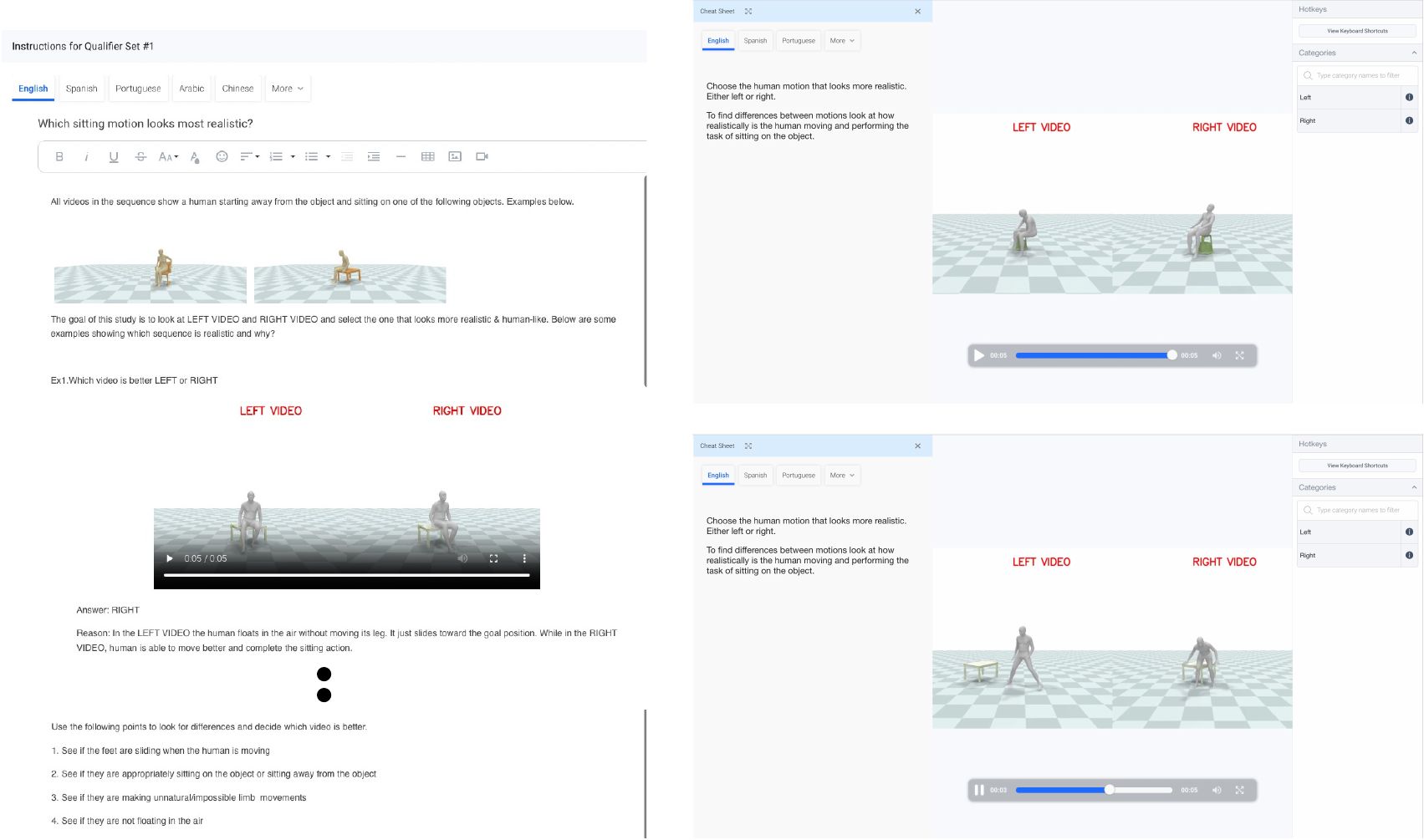} 
    \end{tabular}
    \end{adjustbox}
    \vspace{-2mm}
    \caption{\textbf{A/B Testing User Study} We use this study to compare the quality of motions generated by different methods by requiring them to generate human-object interaction motions. On the left, we show the instruction set following which all users are required to pass a qualification exam to participate in the study. On the right, we show the user interface as visible to users. The users answer the question "Which motion is more realistic" and are required to choose one between ``LEFT VIDEO" or ``RIGHT VIDEO".
    }
    \label{fig:ab_instructions}
\end{figure}
\subsection{A/B Test User Study}
\label{subsec:userstudy-details}
We conduct a user study to qualitatively evaluate the performance of two methods. We design a study such that, given a pair of motions, a user must choose one that is the most realistic. Specifically, we ask the user ``Which motion among the both is more realistic?" when we show them two videos (each containing a motion generated by a different method) ``LEFT VIDEO" \& ``RIGHT VIDEO". \cref{fig:ab_instructions} shows the instructions and user interface from the study. 
We conduct 3 such studies using \texttt{hive.ai}~\cite{hive}, the results of which are in Fig 5 of the main paper.

\label{subsec:userstudy}
\parnobf{Filtering Unreliable Users} We require users to understand instructions given in English. User selection for the study is conditioned on the performance of a qualification test. Users with an accuracy of $\ge$  80\% on this test are allowed to take the study. To ensure continued reliability during the labeling process we randomly mix the real task data with ``obvious" honeypot data where the labels are objective. We require users to have a performance of $\ge$ 89\% on these honeypot tasks. A drop in performance below this results in the user being disqualified from taking the study further.

\subsection{Metrics}
\begin{figure}[h!]
    \centering
    \noindent
    \begin{adjustbox}{max width=0.45\linewidth}
    \begin{tabular}{c}
    \noindent\includegraphics{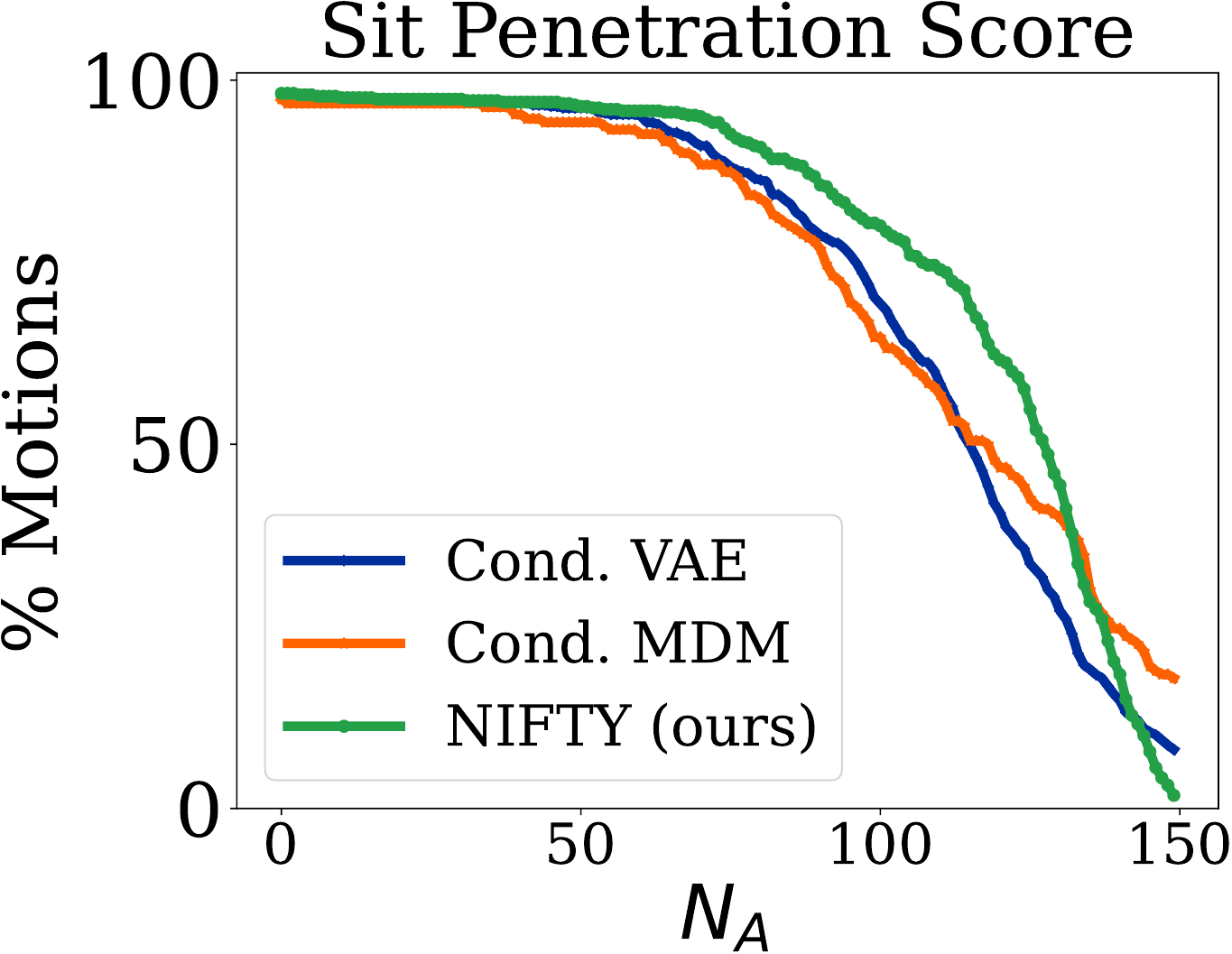} 
    \end{tabular}
    \end{adjustbox}
    \vspace{-2mm}
    \caption{\textbf{Penetration Score Sitting}. We graph the percentage of motion sequences with a penetration score of less than or equal to 2cm (Y-axis), compared to the number of approach frames, denoted as $N_{A}$ (X-axis). Our findings reveal that regardless of the value of $N_{A}$, \ours (\textcolor{ibm_green}{green})  consistently exhibits a greater proportion of motion sequences with low penetration scores.
    }
    \label{fig:pen_sit}
\end{figure}
\begin{figure}[h!]
    \centering
    \noindent
    \begin{adjustbox}{max width=0.45\linewidth}
    \begin{tabular}{c}
    \noindent\includegraphics{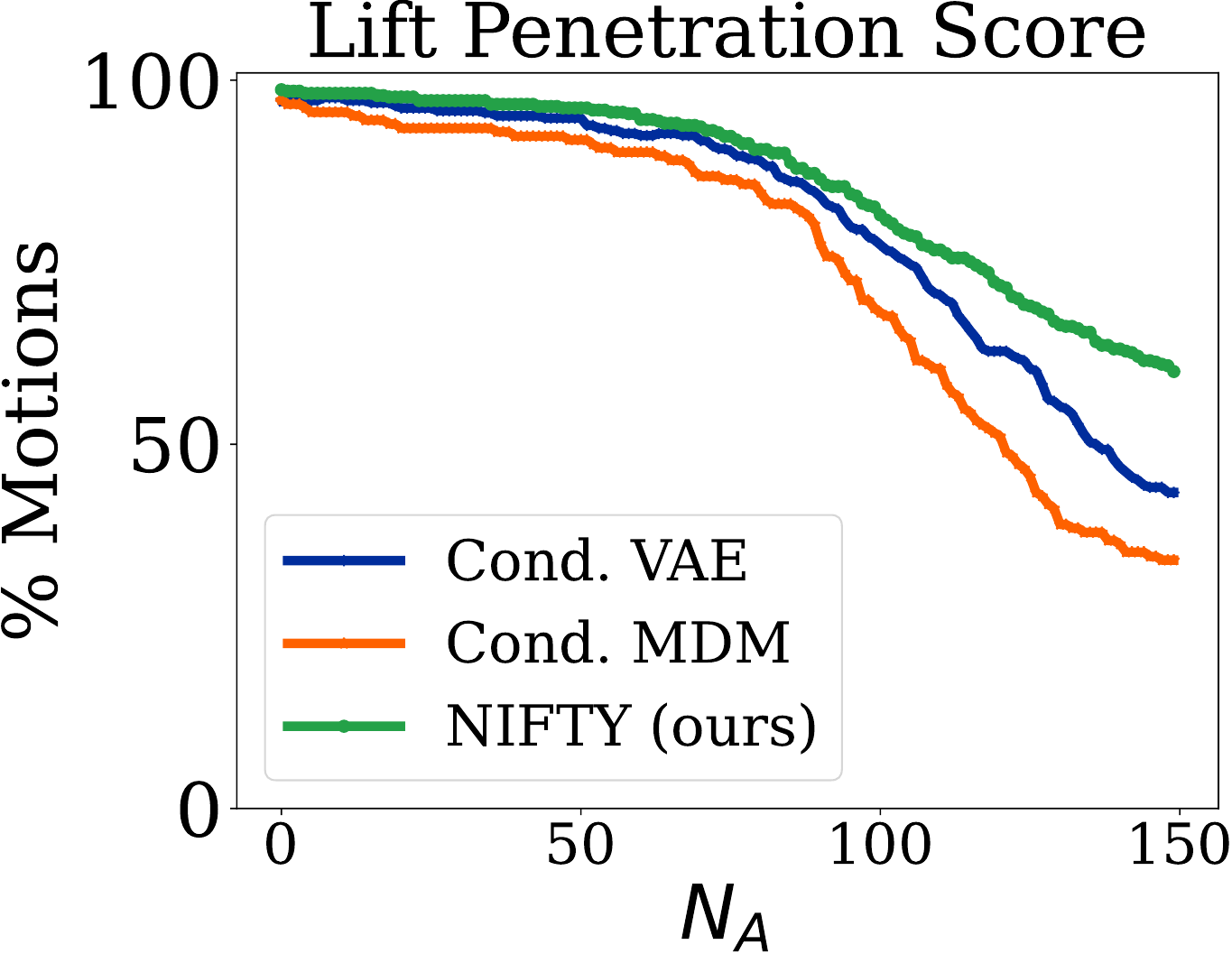} 
    \end{tabular}
    \end{adjustbox}
    \vspace{-2mm}
    \caption{\textbf{Penetration Score Lifting}. We graph the percentage of motion sequences with a penetration score of less than or equal to 2cm (Y-axis), compared to the number of approach frames, denoted as $N_{A}$ (X-axis). Our findings reveal that regardless of the value of $N_{A}$, \ours (\textcolor{ibm_green}{green}) consistently exhibits a greater proportion of motion sequences with low penetration scores.
    }
    \label{fig:pen_lift}
\end{figure}
Apart from performing the user study described in \S\ref{subsec:userstudy} we also evaluate all our models and baselines on several quantitative metrics. We detail these metrics below (apart from the details already described in Sec 4.2 of the main paper).

\parnobf{Penetration Score} To assess the realism of human motion when interacting with an object, we calculate the penetration score during the approach phase. We define the approach phase as the initial $N_{A}$ motion frames from a sequence of 150 frames (5 sec). Our rationale for selecting $N_{A}$ is that during the approach phase, there should be minimal penetration of the human motion into the object geometry. However, during the interaction, there should be increasing contact with the object. These contacts typically result in zero or positive values in the signed distance function (SDF), indicating penetration of points on the object surface into the human SMPL mesh.

We compute $N_{A}$ for sitting and lifting separately based on our synthetic dataset. 
In particular, we determine the first frame index of motion where object penetration distance continues to only \textit{increase} thereafter. We assume that after this point, the person is actually interacting with the object and not just approaching it.  
For sitting, the typical onset of motion interaction occurs after the initial 117 frames of approach, based on the median $N_{A}$. Likewise, lifting has a 15th percentile $N_{A}$ of 124 frames. We use the 15th percentile instead of the median (148 frames) to make this metric more meaningful as 148 frames is almost the end of the complete motion and we wish to evaluate the approach. This difference between {\it sit} and {\it lift} action is due to the difference in their inherent interaction with the object.


For completeness, we also report this performance as a function of different $N_{A}$ values in \cref{fig:pen_sit} (sit) and \cref{fig:pen_lift} (lift).


\parnobf{Skeleton Distance} This metric uses the anchor poses from our human-object interaction data to evaluate whether generated motions faithfully reflect interactions from data. 
We compute a sum over the per-joint location error ($22$ joints in our case) between the final generated interaction pose and the nearest neighbor anchor pose from the training dataset in the joint locations space. We report the average of this metric across generated motions.


\section{Supplemental Results}
\label{sec:newresults}
In this section, we include supplemental analyses to support the evaluations in the main paper that were not included due to space constraints.
First, we evaluate the effect of having a parametric {\it vs} a non-parametric  guidance field in \S\ref{sss:nn}. In \S\ref{sss:samples}, \ref{sss:anchors}, and \ref{sss:ociinp} we evaluate the impact of hyperparameters like the number of samples at inference, number of anchor poses at training, and a variant of our {\it Object Interaction Field} that guides a motion \textit{sequence} instead of just the final {\it interaction frame}. 
We also evaluate the difference in performance across different objects. 

\subsection{Non-Parametric Object Interaction Field}
\label{sss:nn}
We conducted a comparison between our method and a variant where we replaced the object interaction field with a non-parametric field implemented using the nearest neighbor measure. Specifically, during the guidance phase, we identified the nearest anchor pose of the object from the training set and used the difference between this pose and the predicted final pose as the correction. This correction was then utilized to define our distance field and guide the diffusion model accordingly. 

\cref{tab:parametric} presents the comparison between this baseline and our method. 
The skeleton distance metric can be sensitive to outliers (\eg, a few generations that are far from the object), so we additionally report \% Skel. Dist. $\le 25cm$ to get a more robust metric.
The results demonstrate that our learning approach offers a significant improvement of at least 20\% in terms of \textit{Skeleton Distance} $\leq 25$ cm, as well as an additional 10\% in terms of \textit{Contact IoU}. The main paper reports results on the Parametric approach as our primary model.

\begin{table}[h!]
    \setlength{\tabcolsep}{3pt}
    \caption{\textbf{Nearest Neighbor Comparison.} We investigate the effect of learning a parametric function for the Interaction field compared to using the nearest neighbor approach (explained in \S~\ref{sss:nn}). Our results demonstrate that guiding the diffusion model with our learned field outperforms using a non-parametric field. Specifically, for the sitting action dataset, our Parametric method surpasses the Non-Parametric method by 0.09 points in Contact IoU and achieves an 18\% improvement in Skel. Dist $\le 25cm$. Similar trends are observed in the lift action dataset.} 
    \centering 
    \vspace{-1mm}
    \begin{adjustbox}{max width=\linewidth}
    
     \begin{tabular}{l r@{~~}r@{~~}rr@{~~}rrr} \toprule
        
                & \multicolumn{7}{r}{\textbf{Sitting}} \\
        Guidance  & Foot                 & \% {\it D2O}          &  {\it D2O}                              & Skel.                &  \% Skel.                   & Contact               & \% Pen.  \\
       Objective         & Skating $\downarrow$ & $\le 2cm$ $\uparrow$  &  $95^\text{th}\tilde{\%}$ $\downarrow$  & Dist. $\downarrow$   & Dist.$\le 25cm$ $\uparrow$ & IoU    $\uparrow$     &  $\le 2cm$ $\uparrow$   \\
                 \midrule   
         Non-Parametric  & 0.44 & 99.80 & 0.00 & 0.31 & 47.01 & 0.45 & 64.67\\
         Parametric & 0.47 & 99.60 & 0.00 & 0.54 & 65.94 & 0.54 & 65.40 \\
        
    \end{tabular}
    \end{adjustbox}
    
    \begin{adjustbox}{max width=\linewidth}
    
     \begin{tabular}{l s@{~~}s@{~~}ss@{~~}sss} \toprule
        
                & \multicolumn{7}{s}{\textbf{Lifting}} \\
        Guidance  & Foot                 & \% {\it D2O}          &  {\it D2O}                              & Skel.                &  \% Skel.                   & Contact               & \% Pen.  \\
        Objective        & Skating $\downarrow$ & $\le 2cm$ $\uparrow$  &  $95^\text{th}\tilde{\%}$ $\downarrow$  & Dist. $\downarrow$   & Dist.$\le 25cm$ $\uparrow$ & IoU    $\uparrow$     &  $\le 2cm$ $\uparrow$   \\
                 \midrule
        Non-Parametric   & 0.32 & 71.12 & 0.07 & 0.52 & 29.88 & 0.11 & 63.02\\
        Parametric & 0.34 & 77.69 & 0.05 & 0.42 & 61.55 & 0.17 & 69.49 \\
        \midrule
    \end{tabular}
    \end{adjustbox}

    \label{tab:parametric}
   
\end{table}

\subsection{Effect of Number of Samples}
In the main paper, we generate 10 guided samples from the diffusion model and use the one with the best guidance score. We investigate the impact of varying these number of samples in \cref{tab:num_reps}. We observe that increasing the number of samples leads to improved performance. Particular improvements occur when transitioning from 1 sample to 5 samples. Since guidance does not always result in perfect samples, drawing a diverse set gives better chance for a high-quality output. Note that drawing additional samples can be done efficiently in parallel.
\label{sss:samples}

\begin{table}[h!]
    \setlength{\tabcolsep}{3pt}
    \caption{\textbf{Number of Samples Analysis.} We study the impact of drawing multiple samples and guiding them. Drawing more samples helps generate better-quality motions.} 
    \centering 
    \vspace{-1mm}
    \begin{adjustbox}{max width=\linewidth}
    
     \begin{tabular}{l r@{~~}r@{~~}rr@{~~}rrr} \toprule
        
                & \multicolumn{7}{r}{\textbf{Sitting}} \\
        \# Samples  & Foot                 & \% {\it D2O}          &  {\it D2O}                              & Skel.                &  \% Skel.                   & Contact               & \% Pen.  \\
                & Skating $\downarrow$ & $\le 2cm$ $\uparrow$  &  $95^\text{th}\tilde{\%}$ $\downarrow$  & Dist. $\downarrow$   & Dist.$\le 25cm$ $\uparrow$ & IoU    $\uparrow$     &  $\le 2cm$ $\uparrow$   \\
                 \midrule   
         1   & 0.66 & 86.25 & 7.36 & 5.72 & 41.83 & 0.40 & 62.59  \\
         2   & 0.56 & 94.62 & 4.29 & 2.36 & 51.20 & 0.47 & 65.47  \\
         5   & 0.47 & 98.81 & 0.00 & 0.67 & 62.55 & 0.51 & 64.72  \\
         10  & 0.47 & 99.60 & 0.00 & 0.54 & 65.94 & 0.54 & 65.40 \\
        
    \end{tabular}
    \end{adjustbox}

    \begin{adjustbox}{max width=\linewidth}
    
     \begin{tabular}{l s@{~~}s@{~~}ss@{~~}sss} \toprule
        
                & \multicolumn{7}{s}{\textbf{Lifting}} \\
        \# Samples  & Foot                 & \% {\it D2O}          &  {\it D2O}                              & Skel.                &  \% Skel.                   & Contact               & \% Pen.  \\
                & Skating $\downarrow$ & $\le 2cm$ $\uparrow$  &  $95^\text{th}\tilde{\%}$ $\downarrow$  & Dist. $\downarrow$   & Dist.$\le 25cm$ $\uparrow$ & IoU    $\uparrow$     &  $\le 2cm$ $\uparrow$   \\
                 \midrule
        1  & 0.36 & 73.11 & 4.84 & 2.21 & 42.03 & 0.14 & 64.58 \\
        2  & 0.35 & 75.70 & 0.08 & 1.17 & 48.80 & 0.14 & 67.37 \\
        5  & 0.34 & 77.29 & 0.06 & 0.59 & 57.57 & 0.17 & 67.53 \\
        10 & 0.34 & 77.69 & 0.05 & 0.42 & 61.55 & 0.17 & 69.49 \\
        \midrule
    \end{tabular}
    \end{adjustbox}

    \label{tab:num_reps}
   
\end{table}
\subsection{Effect of Number of Anchors Poses}

We also train our  Interaction Field (IF) using subsets of motion that yield a limited number of anchor poses. Specifically, we train the IF using 10\%, 25\%, and 50\% of the available seed anchor poses and report results in \cref{tab:num_anchors}. It is worth noting that \textit{Contact IoU} and \textit{Skeleton Dist} metrics are calculated using all anchor poses in the training set. However, methods trained with only $X$\% of the anchor data will not be able to generate the complete range of seed poses. Therefore, when comparing methods trained with different percentages of seed anchor poses, we primarily assess them based on other metrics, but \textit{Contact IoU} and \textit{Skeleton Dist} are still included for completeness. 

\ours's performance remains stable even with the limited availability of anchor poses. Looking at \textit{Foot Skating}, \textit{D2O}, and \textit{Penetration} metrics, there is not a significant decline in performance. The main paper reports results on 100\% data for \ours.

\label{sss:anchors}

\begin{table}[h!]
    \setlength{\tabcolsep}{3pt}
    \caption{\textbf{Number of Anchors at Training.} We vary the number anchor poses available for training the Interaction Field. We see metrics like  Foot Skating, D20, and Pen. are relatively stable as compared to a number of anchors. The evaluation using Skel.Distance and Contact IoU uses all the anchor poses in the training dataset and this evaluation hence hurts the methods that have access to the less anchor poses during training. For this particular ablation we consider Foot Skating, D2O, and Pen. are primary metrics for this ablation.} 
    \centering 
    \vspace{-1mm}
    \begin{adjustbox}{max width=\linewidth}
    
     \begin{tabular}{l r@{~~}r@{~~}rr@{~~}|rrr} \toprule
        
                & \multicolumn{7}{r}{\textbf{Sitting}} \\
        \% Anchors  & Foot  & \% {\it D2O}   &  {\it D2O}   & \% Pen.  & Skel.   &  \% Skel.    & Contact   \\
                   & Skating $\downarrow$ & $\le 2cm$ $\uparrow$  &  $95^\text{th}\tilde{\%}$ $\downarrow$ &  $\le 2cm$ $\uparrow$ & Dist. $\downarrow$   & Dist.$\le 25cm$ $\uparrow$ & IoU    $\uparrow$  \\
                 \midrule   
         10\%  & 0.55 & 95.82 & 0.00 & 53.02 & 1.90 & 12.35 & 0.27  \\
         25 \%  & 0.54 & 98.01 & 0.00 & 53.86  & 1.28 & 28.88 & 0.34  \\
         50 \%   & 0.49 & 98.21 & 0.00 & 59.23 & 0.96 & 34.86 & 0.40   \\
         100\%   & 0.47 & 99.60 & 0.00 & 65.40 & 0.54 & 65.94 & 0.54  \\
        
    \end{tabular}
    \end{adjustbox}

    \begin{adjustbox}{max width=\linewidth}
    
     \begin{tabular}{l s@{~~}s@{~~}ss@{~~}|sss} \toprule
                & \multicolumn{7}{s}{\textbf{Lifting}} \\
        \% Anchors  & Foot  & \% {\it D2O}   &  {\it D2O}   & \% Pen.  & Skel.   &  \% Skel.    & Contact   \\
                   & Skating $\downarrow$ & $\le 2cm$ $\uparrow$  &  $95^\text{th}\tilde{\%}$ $\downarrow$ &  $\le 2cm$ $\uparrow$ & Dist. $\downarrow$   & Dist.$\le 25cm$ $\uparrow$ & IoU    $\uparrow$  \\
                 \midrule
        10\%  & 0.37 & 83.27 & 0.07 & 50.72 & 0.98 & 14.54 & 0.06  \\
        25\%  & 0.37 & 84.86 & 0.05 & 46.24 & 1.32 & 22.11 & 0.07  \\
        50\%  & 0.36 & 78.49 & 0.06 & 56.34 & 1.01 & 24.90 & 0.08  \\
        100\% & 0.34 & 77.69 & 0.05 & 69.49 & 0.42 & 61.55 & 0.17  \\
        \midrule
    \end{tabular}
    \end{adjustbox}

    \label{tab:num_anchors}
   
\end{table}

\subsection{Effect of Number of Input Frames on Interaction Field}
In the main paper, our interaction field only considers the last interaction pose, denoted as $\tilde{X}$. However, we want to investigate the impact of extending the interaction field to operate on a sequence of frames rather than just the final interaction frame. To achieve this, we modify our Object Interaction Field to process a sequence of frames from $N-m$ to $N$, represented as $\{\tilde{X}_{N-m}\dots\tilde{X}_{N}\}$. Using a transformer encoder, we encode this sequence and obtain a correction vector, denoted as $\Delta\{\tilde{X_{N-m}}\dots \tilde{X_{N}}\}$. In \cref{tab:num_inp_frames}, we present preliminary results using this spatiotemporal configuration. The results indicate that training such an interaction field is feasible but requires a more careful tuning of different hyperparameters, \eg, the guidance weights. Further investigation into this matter is left for future research.
\label{sss:ociinp}

\begin{table}[h!]
    \setlength{\tabcolsep}{3pt}
    \caption{\textbf{Multiple Input Frames to Interaction Field} We show preliminary results on training an interaction field that considers multiple frames as input instead of a single frame like in the main paper. Our results indicate training such a field is feasible the requires further analysis to understand the effect of different hyperparameters.} 
    \centering 
    \vspace{-1mm}
    \begin{adjustbox}{max width=\linewidth}
    
     \begin{tabular}{l r@{~~}r@{~~}rr@{~~}rrr} \toprule
        
                & \multicolumn{7}{r}{\textbf{Sitting}} \\
        \# Input  & Foot                 & \% {\it D2O}          &  {\it D2O}                              & Skel.                &  \% Skel.                   & Contact               & \% Pen.  \\
        Frames        & Skating $\downarrow$ & $\le 2cm$ $\uparrow$  &  $95^\text{th}\tilde{\%}$ $\downarrow$  & Dist. $\downarrow$   & Dist.$\le 25cm$ $\uparrow$ & IoU    $\uparrow$     &  $\le 2cm$ $\uparrow$   \\
                 \midrule   
         1  & 0.47 & 99.60 & 0.00 & 0.54 & 65.94 & 0.54 & 65.40 \\
         5  & 0.66 & 86.25 & 7.36 & 5.72 & 41.83 & 0.40 & 62.59  \\
         10  & 0.56 & 94.62 & 4.29 & 2.36 & 51.20 & 0.47 & 65.47  \\
         15   & 0.47 & 98.81 & 0.00 & 0.67 & 62.55 & 0.51 & 64.72  \\

    \end{tabular}
    \end{adjustbox}

    \begin{adjustbox}{max width=\linewidth}
    
     \begin{tabular}{l s@{~~}s@{~~}ss@{~~}sss} \toprule
        
                & \multicolumn{7}{s}{\textbf{Lifting}} \\
        \# Input  & Foot                 & \% {\it D2O}          &  {\it D2O}                              & Skel.                &  \% Skel.                   & Contact               & \% Pen.  \\
        Frames        & Skating $\downarrow$ & $\le 2cm$ $\uparrow$  &  $95^\text{th}\tilde{\%}$ $\downarrow$  & Dist. $\downarrow$   & Dist.$\le 25cm$ $\uparrow$ & IoU    $\uparrow$     &  $\le 2cm$ $\uparrow$   \\
                 \midrule
        1  & 0.34 & 77.69 & 0.05 & 0.42 & 61.55 & 0.17 & 69.49 \\
        5  & 0.35 & 76.10 & 0.06 & 0.37 & 62.55 & 0.16 & 67.28 \\
        10 & 0.34 & 78.09 & 0.05 & 0.46 & 62.55 & 0.17 & 69.64 \\
        15 & 0.34 & 77.49 & 0.06 & 0.36 & 62.95 & 0.16 & 68.64 \\
        \midrule
    \end{tabular}
    \end{adjustbox}

    \label{tab:num_inp_frames}
   
\end{table}

\subsection{Effect of training Interaction Field in the Local Human Frame}
Our interaction field is object-centric since it takes in a canonical object point cloud as input. 
To test this design choice, we implement the object interaction field in the local frame of the human requiring it to understand the spatial positioning of the object w.r.t to the human motion. 
As shown in \cref{tab:scene}, this leads to a subpar performance across the board on sit and lift actions.
\label{sss:scene}

\begin{table}[h!]
    \setlength{\tabcolsep}{3pt}
    \caption{\textbf{Canonical {\it vs.} Local Human Frame for Interaction Field Training}. We show that training an Interaction Field in the local human motion frame leads to poor performance as comared to } 
    \centering 
    \vspace{-1mm}
    \begin{adjustbox}{max width=\linewidth}
    
     \begin{tabular}{l r@{~~}r@{~~}rr@{~~}rrr} \toprule
        
                & \multicolumn{7}{r}{\textbf{Sitting}} \\
        Interaction  & Foot                 & \% {\it D2O}          &  {\it D2O}                              & Skel.                &  \% Skel.                   & Contact               & \% Pen.  \\
       Field Frame         & Skating $\downarrow$ & $\le 2cm$ $\uparrow$  &  $95^\text{th}\tilde{\%}$ $\downarrow$  & Dist. $\downarrow$   & Dist.$\le 25cm$ $\uparrow$ & IoU    $\uparrow$     &  $\le 2cm$ $\uparrow$   \\
                 \midrule   
         Local Human  & 0.36 & 40.04 & 0.86 & 2.62 & 0.20 & 0.04 & 53.73\\
         Canonical    & 0.47 & 99.60 & 0.00 & 0.54 & 65.94 & 0.54 & 65.40 \\
        
    \end{tabular}
    \end{adjustbox}
    
    \begin{adjustbox}{max width=\linewidth}
    
     \begin{tabular}{l s@{~~}s@{~~}ss@{~~}sss} \toprule
        
                & \multicolumn{7}{s}{\textbf{Lifting}} \\
        Interaction  & Foot                 & \% {\it D2O}          &  {\it D2O}                              & Skel.                &  \% Skel.                   & Contact               & \% Pen.  \\
        Field Frame        & Skating $\downarrow$ & $\le 2cm$ $\uparrow$  &  $95^\text{th}\tilde{\%}$ $\downarrow$  & Dist. $\downarrow$   & Dist.$\le 25cm$ $\uparrow$ & IoU    $\uparrow$     &  $\le 2cm$ $\uparrow$   \\
                 \midrule
        Local Human   & 0.28 & 41.83 & 1.02 & 2.44 & 0.60 & 0.02 & 43.33\\
        Canonical  & 0.34 & 77.69 & 0.05 & 0.42 & 61.55 & 0.17 & 69.49 \\
        \midrule
    \end{tabular}
    \end{adjustbox}

    \label{tab:scene}
   
\end{table}

\subsection{Performance Breakdown Per-Object}
We analyze if the performance of our method is biased towards certain objects by computing the metrics for about 100 interaction motion samples per object instance. 
We show the results of this in~\cref{tab:category}. 
Results indicate that the performance of our method is not dependent on the kind of the object. For instance, in the case of sitting, the performance for sitting on a ``Armchair" {\it vs} ``Chair" are close. This demonstrates the flexibility of the \ours pipeline to a diverse set of objects. 
\label{sss:category}

\begin{table}[h!]
    \setlength{\tabcolsep}{3pt}
    \caption{\textbf{Performance on actions across objects}. We see that \ours's performance is stable across object categories and the framework handles different objects effectively. For instance, the performance on the Armchair and Chair on sitting action are close signaling the flexibility of \ours pipeline.} 
    \centering 
    \vspace{-1mm}
    \begin{adjustbox}{max width=\linewidth}
    
     \begin{tabular}{l r@{~~}r@{~~}rr@{~~}rrr} \toprule
        
                & \multicolumn{7}{r}{\textbf{Sitting}} \\
        Object{~~~~~~~~~~}  & Foot                 & \% {\it D2O}          &  {\it D2O}                              & Skel.                &  \% Skel.                   & Contact               & \% Pen.  \\
                & Skating $\downarrow$ & $\le 2cm$ $\uparrow$  &  $95^\text{th}\tilde{\%}$ $\downarrow$  & Dist. $\downarrow$   & Dist.$\le 25cm$ $\uparrow$ & IoU    $\uparrow$     &  $\le 2cm$ $\uparrow$   \\
                 \midrule   
        Armchair  & 0.42 & 99.05 & 0.00 & 0.42 & 90.48 & 0.44 & 56.73 \\
        Chair & 0.51 & 100.00 & 0.00 & 0.17 & 84.31 & 0.60 & 49.02 \\
        Stool & 0.50 & 96.59 & 0.01 & 0.21 & 68.18 & 0.54 & 72.94 \\
        Table & 0.46 & 100.00 & 0.00 & 0.28 & 55.88 & 0.50 & 68.63 \\
        Yoga Ball & 0.53 & 100.00 & 0.00 & 0.22 & 73.33 & 0.58 & 52.38 \\
        
    \end{tabular}
    \end{adjustbox}

    \begin{adjustbox}{max width=\linewidth}
    
     \begin{tabular}{l s@{~~}s@{~~}ss@{~~}sss} \toprule
                & \multicolumn{7}{s}{\textbf{Lifting}} \\
        Object{~~~}\quad\quad\quad   & Foot                 & \% {\it D2O}          &  {\it D2O}                              & Skel.                &  \% Skel.                   & Contact               & \% Pen.  \\
                & Skating $\downarrow$ & $\le 2cm$ $\uparrow$  &  $95^\text{th}\tilde{\%}$ $\downarrow$  & Dist. $\downarrow$   & Dist.$\le 25cm$ $\uparrow$ & IoU    $\uparrow$     &  $\le 2cm$ $\uparrow$   \\
                 \midrule
        Chair & 0.34 & 86.82 & 0.04 & 0.38 & 70.54 & 0.17 & 59.82 \\
        Stool & 0.36 & 77.24 & 0.06 & 0.24 & 65.04 & 0.13 & 76.84 \\
        Suitcase & 0.33 & 63.85 & 0.06 & 0.20 & 71.54 & 0.28 & 65.06 \\
        Table & 0.29 & 90.00 & 0.03 & 0.64 & 51.67 & 0.15 & 57.41 \\
        \midrule
    \end{tabular}
    \end{adjustbox}
  
    \label{tab:category}
   
\end{table}

\section{Qualitative Results}
\begin{figure}[h!]
    \centering
    \noindent
    \begin{adjustbox}{max width=0.9\linewidth}
    \begin{tabular}{c}
    \noindent\includegraphics{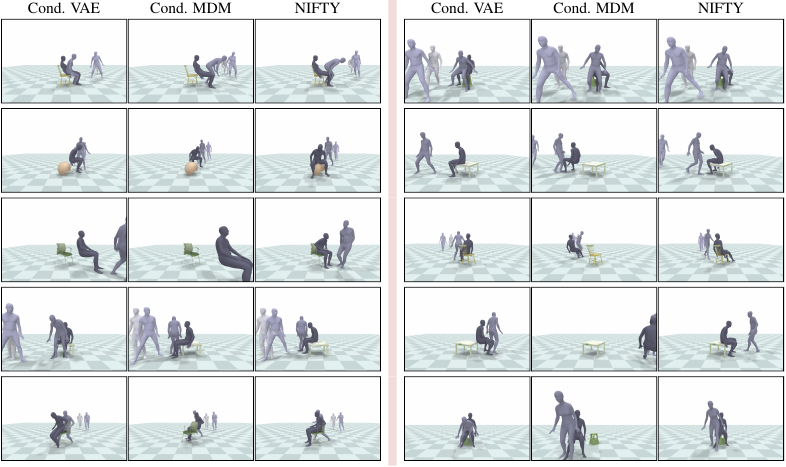} 
    \end{tabular}
    \end{adjustbox}
    \vspace{-2mm}
    \caption{\textbf{Comparison Qualitative Motions Sitting}. Compared to other baselines, our method (\ours) produces more realistic motions. When examining the motion examples generated by the baselines, we notice that in all cases where a person approaches an object to sit, either the person completely misses the object or the sitting pose is not compatible with the object. To better evaluate these results, please refer to the qualitative videos of these motions in the \href{https://nileshkulkarni.github.io/nifty/supplementary.html}{\texttt{supplementary.html}}.
    }
    \label{fig:qual_sit}
\end{figure}

\begin{figure}[h!]
    \centering
    \begin{adjustbox}{max width=0.9\linewidth}
    \includegraphics{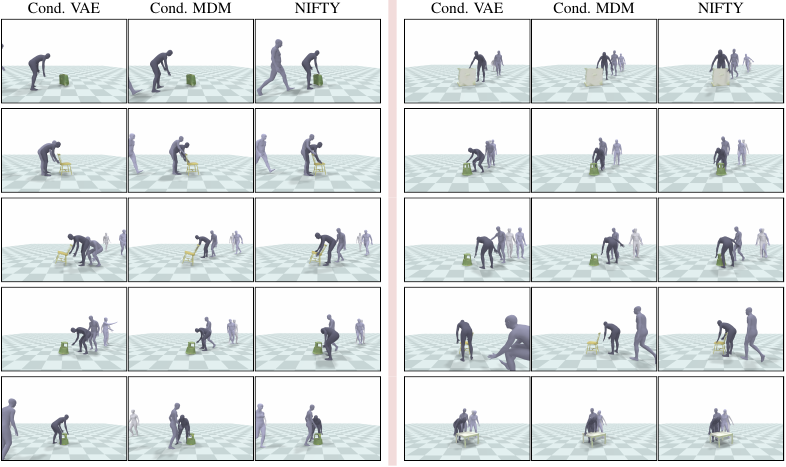} 
    \end{adjustbox}
    \caption{\textbf{Comparison Qualitative Motions Lifting}. \ours generates more realistic motions as compared to the baseline methods. With motions generated using the baseline methods, we see that the lifting stance is often taken far from the object. To better evaluate these results, please refer to the qualitative videos of these motions in the \href{https://nileshkulkarni.github.io/nifty/supplementary.html}{\texttt{supplementary.html}} file.
    }
    \label{fig:qual_lift}
\end{figure}

Motion generation results are best seen as videos on the attached webpage. 
We also include static visualizations here in \cref{fig:qual_sit} and \cref{fig:qual_lift}. The webpage additionally also shows visualizations (~10 motions) from our method for every object in our dataset.

\label{sec:qualitative}

\section{Limitations}
\label{sec:failures}

Our proposed pipeline demonstrates the ability to achieve human-object interaction results with a  diverse sets of objects while only relying on a limited number of anchor poses. One of the key factors contributing to the performance of \ours is the utilization of a pretrained motion model~\cite{rempe2021humor} trained on the AMASS repository~\cite{mahmood2019amass}. Our data generation pipeline has the capability to generate motions and interpolate between existing data in this dataset. However, in cases where a completely novel and extreme seed anchor pose is provided, such as a headstand, HuMoR would struggle to generate reasonable and high-quality motion sequences. Developing more robust motion models which can handle such poses, would be beneficial.

Furthermore, during the inference stage, it is necessary to draw multiple samples from the diffusion model and guide them. This approach yields significantly better performance compared to guiding only a single sample. Exploring research directions that can enhance the stability of the guidance process would be valuable in consistently generating high-quality interaction motions.

\end{document}